\documentclass{article} 
\usepackage{iclr2027_conference,times}

\usepackage{amsmath,amsfonts,bm}

\def\eqref#1{equation~\ref{#1}}

\def\1{\bm{1}}

\DeclareMathAlphabet{\mathsfit}{\encodingdefault}{\sfdefault}{m}{sl}
\SetMathAlphabet{\mathsfit}{bold}{\encodingdefault}{\sfdefault}{bx}{n}

\usepackage{hyperref}
\usepackage{url}
\usepackage{graphicx}
\usepackage{algorithm}
\usepackage{algorithmic}
\usepackage{placeins}
\usepackage{enumitem}
\usepackage{subcaption}

\usepackage{listings}
\usepackage{xcolor}
\definecolor{codegreen}{rgb}{0,0.6,0}
\definecolor{codegray}{rgb}{0.5,0.5,0.5}
\definecolor{codepurple}{rgb}{0.58,0,0.82}
\definecolor{backcolour}{rgb}{0.95,0.95,0.92}
\definecolor{result_color}{rgb}{0.8,0.95,0.8}

\lstdefinestyle{mystyle}{
    backgroundcolor=\color{backcolour},   
    commentstyle=\color{codegreen},
    keywordstyle=\color{magenta},
    numberstyle=\tiny\color{codegray},
    stringstyle=\color{codepurple},
    basicstyle=\ttfamily,
    breakatwhitespace=false,         
    breaklines=true,                 
    captionpos=b,                    
    keepspaces=true,                 
    numbers=left,                    
    numbersep=5pt,                  
    showspaces=false,                
    showstringspaces=false,
    showtabs=false,                  
    tabsize=2
}
\lstdefinestyle{terminal}{
    basicstyle=\ttfamily\color{black}\small,
    backgroundcolor=\color{gray!10},
    frame=none, 
    framerule=1pt, 
    framexleftmargin=0pt, 
    breaklines=true,
    breakatwhitespace=true,
    numbers=none, 
    keepspaces=false, 
    tabsize=2,
}

\usepackage{arydshln}
\usepackage{listings}
\usepackage{rotating}
\usepackage{afterpage}
\usepackage{longtable}
\usepackage{amsmath}
\usepackage{booktabs}
\usepackage{xcolor}
\usepackage{colortbl}
\usepackage{multirow}
\usepackage{amssymb}

\definecolor{highperf}{RGB}{0, 102, 51}      
\definecolor{medperf}{RGB}{255, 204, 0}      
\definecolor{lowperf}{RGB}{204, 0, 0}        
\definecolor{neutralperf}{RGB}{220, 220, 220} 

\usepackage{tikz}
\usetikzlibrary{tikzmark}
\newcommand{\hc}[1]{
  \cellcolor{highperf!30}#1}
\newcommand{\mc}[1]{
  \cellcolor{medperf!40}#1}
\newcommand{\lc}[1]{
  \cellcolor{lowperf!40}#1}
\newcommand{\nc}[1]{
  \cellcolor{neutralperf!50}#1}

\definecolor{highmidperf}{RGB}{194,232,196} 
\definecolor{lowmidperf}{RGB}{255, 153, 153} 

\newcommand{\hmc}[1]{
  \cellcolor{highmidperf!35}#1}
\newcommand{\lmc}[1]{
  \cellcolor{lowmidperf!45}#1}

\newcommand{\hmcOne}[1]{%
  \cellcolor{highperf!20}#1}
  
\newcommand{\hmcThree}[1]{
  \cellcolor{highperf!48}#1}

\newcommand{\msd}[2]{\ensuremath{#1{\scriptstyle\,\pm\,#2}}}
\newcommand{\bmsd}[2]{\ensuremath{\mathbf{#1}{\scriptstyle\,\pm\,#2}}}

\usepackage{xcolor}
\usetikzlibrary{fit,calc}

\newcommand{\marked}[2][]{%
  \tikz[remember picture, baseline=(cell.base)] {
    \node[inner sep=0pt, outer sep=0pt] (cell) {#2};
    \ifx&#1&\else
      \coordinate (#1) at (cell.center);
    \fi
  }%
}

\title{ORBIT: Offset-based Activation Steering for Cross-lingual Multimodal Fact Verification}

\author{
  Huy Trieu\textsuperscript{1,2} \hspace{0.25cm} 
  Duy-Minh Nguyen-Tran\textsuperscript{3} \hspace{0.25cm} 
  Thanh-Dat Nguyen\textsuperscript{4}\\[0.25cm] \hspace{0.005cm}
  \textbf{Huy Tien Nguyen\textsuperscript{1,2} \hspace{0.25cm} 
  Tung Le\textsuperscript{1,2,}}\thanks{Corresponding author: lttung@fit.hcmus.edu.vn}\\[0.25cm]
  \textsuperscript{1}Faculty of Information Technology, University of Science, Ho Chi Minh city, Vietnam \\
  \textsuperscript{2}Vietnam National University, Ho Chi Minh city, Vietnam\\
  \textsuperscript{3}Japan Advanced Institute of Science and Technology, Ishikawa, Japan\\
  \textsuperscript{4}California State Polytechnic University, Pomona, USA
}

\iclrfinalcopy 
\begin{document}

\maketitle

\begin{abstract}
Multimodal fact verification faces a fundamental accessibility bottleneck: while large proprietary models exhibit cross-lingual robustness, small vision-language models suffer a severe, disproportionate accuracy collapse on low-resource languages. Closing this gap traditionally requires prohibitive scale or extensive language-specific fine-tuning. In this work, we propose Offset Recovery via Bilingual Inference-time Technique (ORBIT), an algorithmic intervention that mitigates this collapse at inference time. ORBIT estimates span-specific offset vectors from parallel claim--evidence instances and applies them to claim and evidence representations, 
targeting the cross-lingual degradation induced by low-resource evidence.
Experiments across 81 configurations reveal that ORBIT substantially recovers both accuracy and F1-score in low-resource settings, exposes calibration offset norms as a diagnostic proxy, and 
outperforms translation-based and steering baselines.
\end{abstract}

\section{Introduction}
\label{sec:introduction}

Misinformation today is rarely confined to a single modality or language. 
A false narrative from a conflict zone may spread with a Spanish caption and reach Vietnamese fact-checkers, but the evidence needed to refute it may exist only in low-resource local reports. Nevertheless, such misinformation often persists in linguistically underserved communities. Multimodal fact verification, which involves assessing a claim against associated text and image evidence, emerges as a crucial defense against such complex content~\citep{yao-etal-2023-mocheg}.

However, current multimodal fact verification benchmarks are predominantly monolingual and English-centric. Datasets such as Factify~\citep{mishra2022factify}, MOCHEG~\citep{yao-etal-2023-mocheg}, VERITE~\citep{papadopoulos2024verite}, and MM-FakeBench~\citep{liu2025mmfakebench} implicitly assume that claims, evidence, and models all operate in the same high-resource language. 
This assumption breaks down in real-world scenarios: a model may verify a Javanese claim against English documents, or assess an English claim using Tigrinya evidence. 
We investigate a fundamental question: \textit{When visual evidence remains constant, how does variation in the language of textual evidence affect verification accuracy?}

Prior work in cross-lingual natural language inference documents the brittleness of text-only models in low-resource scenarios~\citep{conneau-etal-2018-xnli, hu-etal-2020-xtreme}. However, these findings do not directly transfer to the multimodal setting. Vision-language models (VLMs) are trained to ground language in visual perception, a process heavily skewed toward English image captions and Western visual contexts~\citep{pmlr-v162-bugliarello22a,thapliyal-etal-2022-crossmodal}. 
Consequently, while visual encoding may remain robust, cross-modal alignment degrades substantially for low-resource language evidence, giving rise to language-conditioned visual grounding errors~\citep{Zhou_2021_CVPR,pfeiffer-etal-2022-xgqa}.

While recent large proprietary models demonstrate improved robustness to cross-lingual setups on multilingual benchmarks~\citep{team2026gemini,openai2026gpt54,anthropic2026claude}, this performance comes at the cost of accessibility and transparency. For fact-checking organizations operating with limited computational resources in linguistically diverse regions, small vision-language models (SVLMs) -- typically under 10B parameters~\citep{AHMED2026103805} -- represent the most viable option for local deployment. 
Despite this practical importance, their cross-lingual behavior remains least understood precisely where SVLMs are most needed. It remains unknown whether SVLMs merely underperform as scaled-down replicas, or suffer from a severe low-resource collapse rooted in the breakdown of cross-modal alignment.

\begin{figure}
    \centerline{\includegraphics[width=0.9\textwidth]{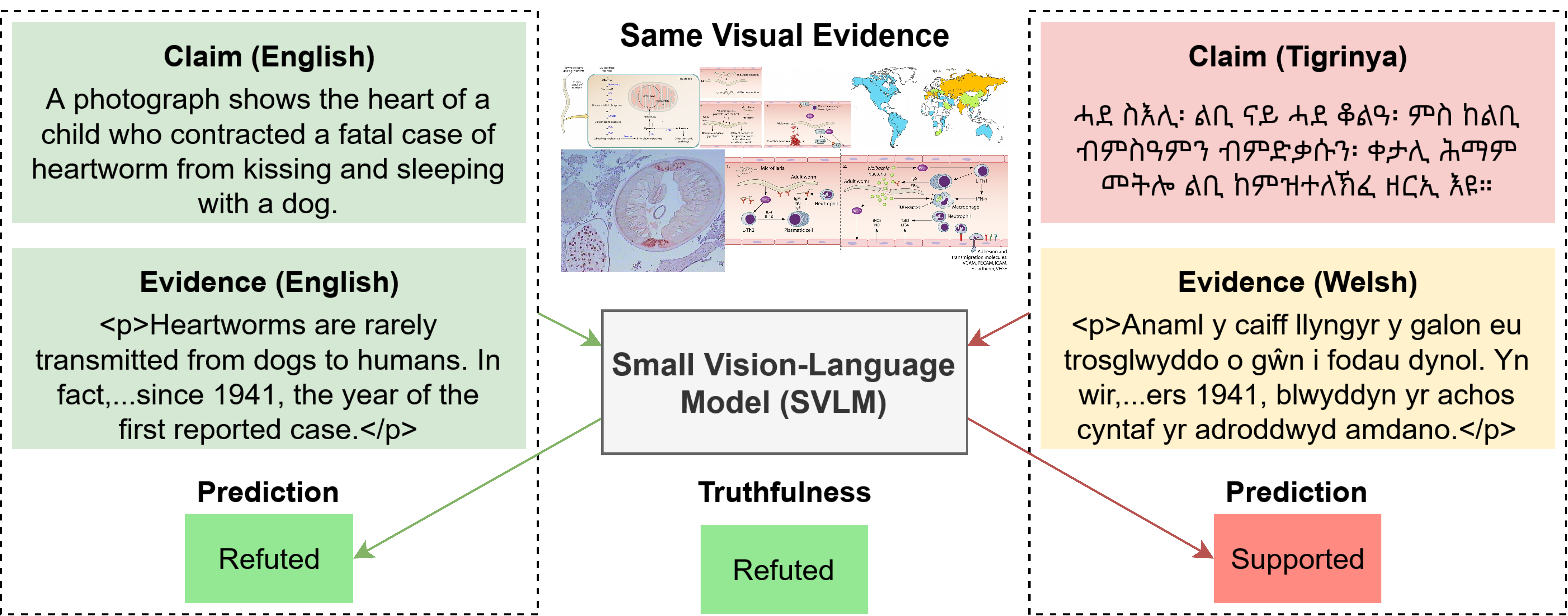}}
    \caption{Overview of cross-lingual multimodal fact verification. Identical visual evidence and semantically equivalent text yield divergent predictions as claim and evidence languages vary.}
    \label{fig:problem}
\end{figure}

To systematically investigate this question, we construct the controlled diagnostic testbed \textbf{MOCHEG-XL} by extending MOCHEG~\citep{yao-etal-2023-mocheg} via machine translation into eight target languages, yielding nine languages including English.
We select MOCHEG because it provides gold textual evidence paired with each claim--image instance, and remains challenging even for strong recent baselines~\citep{chen-etal-2024-metasumperceiver,lee-etal-2024-train,pmlr-v267-braun25b,Pang_Li_Zhang_Wang_Zhang_2025}.
We emphasize that this dataset is designed as an \textbf{analytical instrument} rather than a human-verified benchmark. By holding visual content constant while varying the language of claims and evidence across 81 distinct configurations, we isolate and quantify performance degradation attributable solely to linguistic shift. 
This controlled setup reveals cross-lingual brittleness that is otherwise confounded by content variation in naturally multilingual data.

Observing a disproportionate and severe collapse in low-resource languages, we introduce \textbf{Offset Recovery via Bilingual Inference-time Technique (ORBIT)}, a lightweight inference-time intervention designed to restore low-resource cross-modal grounding. 
ORBIT estimates span-specific offset vectors from parallel claim--evidence instances and applies them separately to claim and evidence representations, directly targeting the multimodal grounding failure induced by low-resource evidence. 
Operating without gradient updates, it avoids catastrophic forgetting and overfitting risks~\citep{Sung_2022_CVPR, vu-etal-2022-overcoming} while bypassing the prohibitive latency of translation-based pipelines.

Our contributions are threefold:
\begin{enumerate}
    \item We conduct a systematic empirical analysis quantifying cross-lingual degradation in SVLMs for multimodal fact verification. We expose a disproportionate and severe collapse in low-resource languages, even when visual content is held constant.
    \item We propose ORBIT, a training-free activation steering method that recovers low-resource performance at modest inference-time cost. We additionally show that calibration offset norms serve as a lightweight diagnostic proxy for predicting cross-lingual collapse without full evaluation.
    \item We will publicly release our translated evaluation configurations (MOCHEG-XL) as a diagnostic toolkit to facilitate future research on cross-lingual robustness of vision-language models.
\end{enumerate}

\section{Related Work}
\label{sec:related_work}

Multimodal fact verification has largely been developed and evaluated under the assumption of monolingual, English-centric settings. Existing benchmarks assume linguistic homogeneity: claims, evidence, and models operate in the same high-resource language. We show that language shift and visual grounding are not independent failures; rather, they compound. Prior work treats cross-lingual transfer and multimodal reasoning as separate problems; consequently, these lines of work cannot diagnose their interaction.

Early multimodal fact verification is largely shaped by binary out-of-context detection benchmarks, with Factify~\citep{mishra2022factify} and NewsCLIPings~\citep{luo-etal-2021-newsclippings} representative of an approach that flags image misuse without requiring evidence-based reasoning about \textit{why} a claim is false. While MOCHEG~\citep{yao-etal-2023-mocheg} and subsequent efforts~\citep{papadopoulos2024verite,liu2025mmfakebench} move toward fine-grained, evidence-grounded verdicts, they preserve the same blind spot of linguistic homogeneity. We isolate the effect of language shift by extending MOCHEG cross-lingually, holding content constant while varying language.

Do cross-lingual gaps persist when images provide a shared grounding signal? Cross-lingual NLP benchmarks~\citep{conneau-etal-2018-xnli,hu-etal-2020-xtreme} reveal systematic performance drops across languages. While recent inference-time methods like INCLINE~\citep{wang-etal-2025-bridging} attempt to map source languages to a target language space using learned alignment matrices, they operate strictly within the text modality. 
Studies on vision-language tasks~\citep{pfeiffer-etal-2022-xgqa,Zhou_2021_CVPR} demonstrate that models suffer language-conditioned grounding errors with non-English text~\citep{thapliyal-etal-2022-crossmodal}. 
These findings, however, are drawn from high-resource multilingual settings and confined to generative or retrieval tasks. Low-resource grounding failures in structured verification tasks remain unexplored.
Fact verification is fundamentally different, as it requires fine-grained tri-modal alignment among claim semantics, evidence text, and image content to output a discrete verdict. We provide the first controlled framework to isolate and quantify cross-lingual grounding failures in multimodal fact verification.

The practical obstacle is that fine-tuning on target-language data is often challenging due to data scarcity. Labeled cross-lingual multimodal data is scarce, and gradient-based updates risk catastrophic forgetting~\citep{Sung_2022_CVPR,vu-etal-2022-overcoming}. Parameter-efficient methods~\citep{pmlr-v97-houlsby19a,hu2022lora,lee-etal-2025-cross} reduce trainable parameters but still require supervision. Activation steering offers a training-free alternative by adding precomputed direction vectors to hidden states during inference. In language models, it improves truthfulness~\citep{li2023inference}; in vision-language models, it mitigates hallucinations~\citep{wu-etal-2025-sharp}; and in cross-lingual settings, approaches like CLAS~\citep{pokharel2026cross} modulate text representations by dynamically scaling shared versus language-specific neurons. 
While most of these approaches assume monolingual or text-only inputs, CLAIM~\citep{ye-etal-2025-claim} extends intervention to multilingual VLMs for object hallucination. However, it requires training per-head SVM probes, steers only the last token's attention, and assumes a single target language rather than misaligned claim--evidence pairs. We bridge this gap with a highly efficient inference-time intervention.
Rather than dynamic neuron-level masking, which requires per-token computation, we estimate a static, per-language correction vector from parallel evidence pairs for each target language. Each vector applies an additive shift to realign that language's representations toward the anchor space, bypassing auxiliary decoding while preserving efficiency.

\section{Cross‑Lingual Collapse}
\label{sec:cross_lingual_collapse}

We isolate linguistic variation from visual content to measure its effect on multimodal fact verification. We use Qwen3-VL 4B\footnote{\href{https://huggingface.co/Qwen/Qwen3-VL-4B-Instruct}{Qwen/Qwen3-VL-4B-Instruct}}~\citep{bai2025qwen3}, the only open-source 4B-parameter SVLM in our candidate pool to exceed 50\% accuracy on MOCHEG, and evaluate nine typologically diverse languages. We treat English (en) as the reference baseline, as it is the source language of MOCHEG~\citep{yao-etal-2023-mocheg}. The remaining eight languages are categorized by pretraining data availability and typological coverage:

\begin{enumerate}
    \item \textbf{High-resource languages:} Spanish (es), Russian (ru), Vietnamese (vi), Japanese (ja).
    \item \textbf{Low-resource languages:} Javanese (jw), Welsh (cy), M\={a}ori (mi), Tigrinya (ti).
\end{enumerate}

\begin{figure}
\centering
\tiny
\setlength{\tabcolsep}{2.75pt}
\setlength{\fboxsep}{1pt}

\begin{minipage}[t]{0.5\textwidth}
\centering
\renewcommand{\arraystretch}{1.15}
\begin{tabular}{l|ccccccccc}
\toprule
\multirow{2}{*}{\textbf{Claim}} & \multicolumn{9}{c}{\textbf{Evidence Language}} \\
\cmidrule{2-10}
& \textbf{en} & \textbf{es} & \textbf{ru} & \textbf{vi} & \textbf{ja} & \textbf{jw} & \textbf{cy} & \textbf{mi} & \textbf{ti} \\
\midrule
\textbf{en} & \marked[acc-tl]{\hc{55.00}} & \hc{53.64} & \hc{52.99} & \hc{51.72} & \marked[acc-tr]{\hc{51.84}} & \mc{47.71} & \mc{43.24} & \lc{38.70} & \lc{36.98} \\
\textbf{es} & \hc{54.55} & \hc{54.01} & \hc{52.54} & \hc{51.19} & \hc{50.53} & \mc{45.74} & \mc{42.51} & \lc{37.59} & \lc{35.54} \\
\textbf{ru} & \hc{54.01} & \hc{52.78} & \hc{53.03} & \hc{50.74} & \mc{49.06} & \mc{44.80} & \mc{41.89} & \lc{38.21} & \lc{35.46} \\
\textbf{vi} & \hc{54.05} & \hc{51.11} & \hc{50.70} & \hc{50.70} & \mc{48.57} & \mc{45.29} & \mc{40.70} & \lc{35.59} & \lc{35.42} \\
\textbf{ja} & \marked[acc-bl]{\hc{54.46}} & \hc{51.43} & \hc{50.04} & \hc{50.29} & \marked[acc-br]{\mc{49.96}} & \mc{44.59} & \mc{40.34} & \lc{37.92} & \lc{35.18} \\
\textbf{jw} & \mc{48.32} & \mc{46.23} & \mc{44.55} & \mc{45.25} & \mc{43.94} & \mc{45.17} & \lc{37.10} & \lc{33.82} & \lc{33.09} \\
\textbf{cy} & \mc{46.40} & \mc{44.51} & \mc{42.14} & \mc{43.20} & \mc{41.85} & \lc{37.22} & \mc{42.51} & \lc{33.78} & \lc{33.33} \\
\textbf{mi} & \mc{42.79} & \mc{41.11} & \mc{40.21} & \lc{39.97} & \lc{39.72} & \lc{36.12} & \lc{34.48} & \lc{35.38} & \lc{32.76} \\
\textbf{ti} & \mc{40.09} & \lc{39.68} & \lc{36.90} & \lc{37.35} & \lc{36.81} & \lc{35.26} & \lc{34.32} & \lc{33.33} & \lc{35.01} \\
\bottomrule
\end{tabular}
\caption*{Accuracy (\%)}
\end{minipage}%
\hfill
%
\begin{minipage}[t]{0.5\textwidth}
\centering
\renewcommand{\arraystretch}{1.15}
\begin{tabular}{l|ccccccccc}
\toprule
\multirow{2}{*}{\textbf{Claim}} & \multicolumn{9}{c}{\textbf{Evidence Language}} \\
\cmidrule{2-10}
& \textbf{en} & \textbf{es} & \textbf{ru} & \textbf{vi} & \textbf{ja} & \textbf{jw} & \textbf{cy} & \textbf{mi} & \textbf{ti} \\
\midrule
\textbf{en} & \marked[f1-tl]{\hc{53.28}} & \hc{51.77} & \hc{51.50} & \hc{50.76} & \marked[f1-tr]{\hc{51.32}} & \mc{48.06} & \mc{43.78} & \lc{38.51} & \lc{35.89} \\
\textbf{es} & \hc{52.17} & \hc{52.24} & \hc{51.04} & \hc{50.25} & \hc{50.11} & \mc{46.11} & \mc{43.04} & \lc{37.46} & \lc{34.15} \\
\textbf{ru} & \hc{51.81} & \hc{51.39} & \hc{51.47} & \mc{49.81} & \mc{48.68} & \mc{45.20} & \mc{42.38} & \lc{38.11} & \lc{34.25} \\
\textbf{vi} & \hc{52.18} & \hc{50.16} & \mc{49.57} & \mc{49.66} & \mc{48.45} & \mc{45.70} & \mc{41.18} & \lc{35.06} & \lc{33.70} \\
\textbf{ja} & \marked[f1-bl]{\hc{52.15}} & \hc{50.08} & \mc{48.94} & \mc{49.49} & \marked[f1-br]{\mc{49.46}} & \mc{44.92} & \mc{40.88} & \lc{37.69} & \lc{33.55} \\
\textbf{jw} & \mc{47.40} & \mc{46.02} & \mc{44.43} & \mc{45.38} & \mc{44.35} & \mc{45.41} & \lc{36.84} & \lc{32.11} & \lc{30.37} \\
\textbf{cy} & \mc{46.18} & \mc{44.82} & \mc{42.40} & \mc{43.66} & \mc{42.38} & \lc{36.50} & \mc{42.53} & \lc{31.29} & \lc{28.87} \\
\textbf{mi} & \mc{43.11} & \mc{41.37} & \mc{40.46} & \mc{40.15} & \lc{39.90} & \lc{35.03} & \lc{32.87} & \lc{35.02} & \lc{27.58} \\
\textbf{ti} & \mc{40.36} & \lc{39.60} & \lc{36.60} & \lc{36.99} & \lc{35.49} & \lc{33.19} & \lc{31.98} & \lc{29.51} & \lc{32.60} \\
\bottomrule
\end{tabular}
\caption*{F1-score (\%)}
\end{minipage}

\begin{tikzpicture}[remember picture, overlay]
  \draw[black, line width=0.45pt] 
    ($(acc-bl)+(-0.34,-0.155)$) -- ($(acc-br)+(0.34,-0.155)$) -- ($(acc-tr)+(0.34,0.1275)$);
    
  \draw[black, line width=0.45pt] 
    ($(f1-bl)+(-0.34,-0.155)$) -- ($(f1-br)+(0.34,-0.155)$) -- ($(f1-tr)+(0.34,0.1275)$);
\end{tikzpicture}

\vspace{0.3cm}

\begin{minipage}{\textwidth}
\centering
\scriptsize
\textbf{Performance levels:} 
\colorbox{highperf!30}{\strut~High ($>$\,50\%)~} 
\colorbox{medperf!40}{\strut~Medium (40--50\%)~} 
\colorbox{lowperf!50}{\strut~Low ($<$\,40\%)~}
\end{minipage}

\caption{Cross-lingual performance of Qwen3-VL across 81 claim--evidence configurations. Low-resource textual evidence (jw, cy, mi, ti) causes severe degradation, with M\={a}ori and Tigrinya near random. Same-language alignment benefit persists only for high-resource languages, which remain stable across both monolingual and cross-lingual settings.}
\label{fig:cross_lingual_heatmap}
\end{figure}

Translations for the eight target languages are generated using the Google Cloud Translation API rather than open-source encoder--decoder models such as NLLB-200~\citep{costa2022nllb}. NLLB-200 is trained with a maximum sequence length of 512 tokens, which MOCHEG evidence passages frequently exceed (see Appendix~\ref{appendix:mocheg_stats}). Prior work establishes that autoregressive neural translation suffers from exposure bias and hallucination as sequence length increases~\citep{wang-sennrich-2020-exposure}. We therefore rely on the Google Cloud Translation API for its greater stability on long-form text.
Translation quality metrics based on LLM-as-a-judge~\citep{NEURIPS2023_91f18a12} evaluation are provided in Appendix~\ref{appendix:translation_quality}.

By holding visual content constant while varying claim and evidence languages across 81 distinct configurations, we isolate performance degradation attributable solely to linguistic shift. 
Figure~\ref{fig:cross_lingual_heatmap} reports accuracy and macro F1-score across all language pairs. The results show that evidence language is the dominant driver of collapse. 
When evidence shifts into a low-resource language, accuracy collapses across every claim language. For instance, while English claims achieve 55.00\% accuracy with English evidence, using Javanese evidence causes a 7.29-point drop. This indicates that visual encoding remains robust, but cross-modal alignment deteriorates when evidence language shifts.

At the extreme end of the resource spectrum, performance collapses entirely. M\={a}ori and Tigrinya evidence reduce the model to near-random guessing: accuracy falls to 33--39\% and F1 to 29--38\%, regardless of claim language. 
This is not mere underperformance; it reflects a severe breakdown in low-resource text comprehension and cross-modal grounding: the model fails to map low-resource textual features into the shared vision-language space.
This breakdown also explains why the same-language advantage disappears under low-resource evidence. Same-language claim--evidence pairs outperform mismatched ones for high-resource languages, yet this diagonal advantage vanishes entirely for Javanese, Welsh, M\={a}ori, and Tigrinya. Without adequate multilingual pretraining signal, the model cannot maintain coherent representations. The effect becomes uniform: the claim language no longer matters, because the evidence encoding has already failed.

These findings expose a fundamental vulnerability: even with identical visual inputs, cross-lingual text understanding and multimodal grounding collapse under low-resource evidence. Although this analysis focuses on Qwen3-VL, we also evaluate Gemma 4 E2B\footnote{\href{https://huggingface.co/google/gemma-4-E2B-it}{google/gemma-4-E2B-it}}~\citep{team2026gemma} and LFM2-VL 3B\footnote{\href{https://huggingface.co/LiquidAI/LFM2-VL-3B}{LiquidAI/LFM2-VL-3B}}~\citep{amini2025lfm2}; their zero-shot results (Appendix~\ref{appendix:additional_base_results}) exhibit the same evidence-language collapse, though the severity varies across models.
We further show in Section~\ref{sec:norm_analysis} that the severity of this collapse can be predicted from calibration offset norms alone, offering a lightweight diagnostic before full evaluation. This divergence motivates targeted, inference-time intervention; the next section details our approach to realigning hidden representations without gradient updates. 

\section{ORBIT}
\label{sec:method}

We propose \textbf{ORBIT}, a training-free intervention that tackles the systematic degradation Section~\ref{sec:cross_lingual_collapse} reveals. By estimating and applying correction vectors to the model's hidden states during inference, this approach realigns low-resource language text representations toward their anchor-language counterparts.

\begin{figure}
    \centerline{\includegraphics[width=1\textwidth]{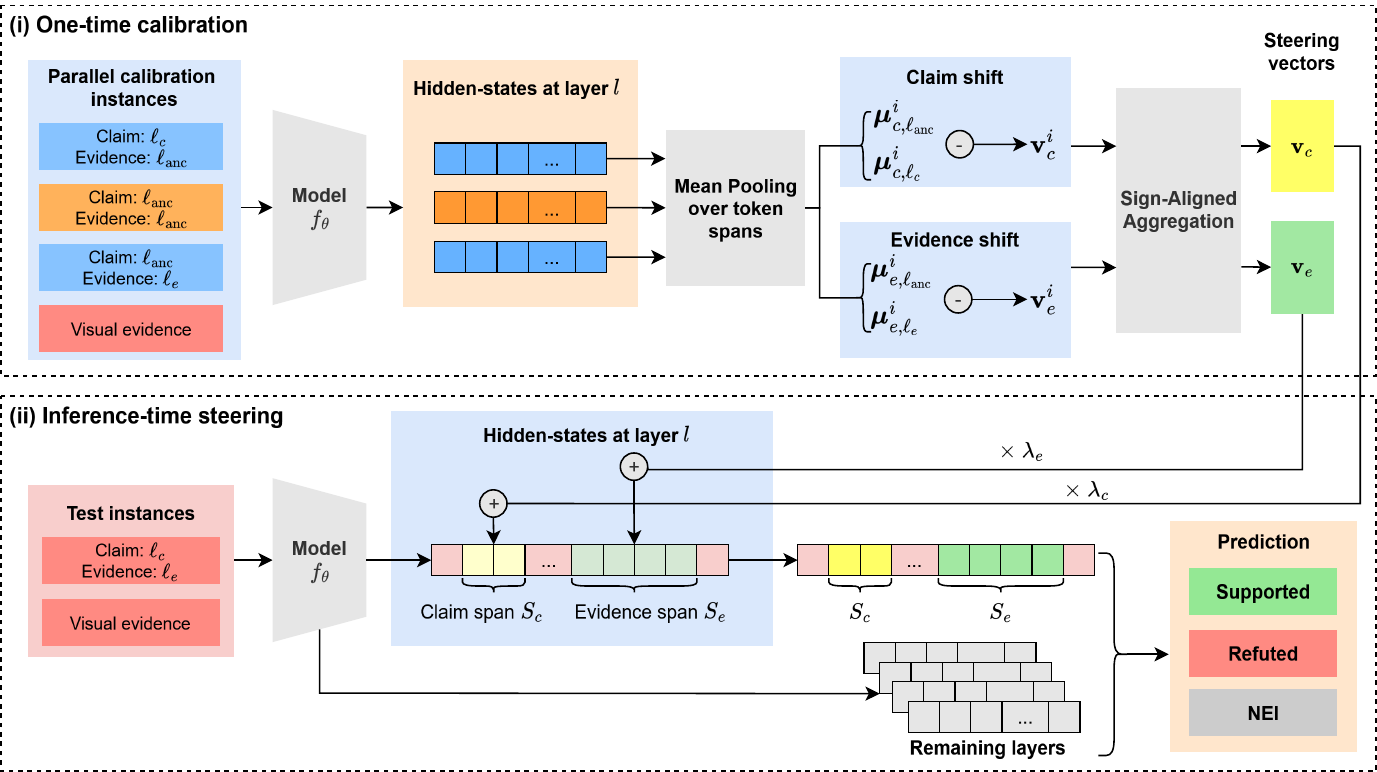}}
    \caption{Overview of ORBIT. (i) One-time calibration aggregates per-sample claim and evidence shifts from parallel instances into per-language steering vectors $\mathbf{v}_c$ and $\mathbf{v}_e$. (ii) Inference-time steering adds these vectors to hidden states at layer $l$, realigning the claim span $S_c$ and evidence span $S_e$ toward the anchor space without parameter updates.}
    \label{fig:architecture}
\end{figure}

Consider a multimodal fact verification instance $(c, e, \mathrm{I})$ where $c$ is the claim in language $\ell_c$, $e$ is the textual evidence in language $\ell_e$, and $\mathrm{I}$ is the visual evidence. ORBIT operates without access to task labels. Let $\mathbf{h}^{(l)}_{\ell_c,\ell_e} \in \mathbb{R}^{T \times d}$ denote the hidden states at layer $l$, where $T$ is the sequence length and $d$ is the hidden dimension.
ORBIT estimates two language-specific correction vectors to realign representations toward an anchor language $\ell_{\text{anc}}$ during inference: $\mathbf{v}_c \in \mathbb{R}^d$ for the claim component and $\mathbf{v}_e \in \mathbb{R}^d$ for the evidence component. These corrections are applied independently without updating model parameters.

\subsection{Estimating Steering Vectors}
\label{sec:steering_vectors}

ORBIT operates in a one-time calibration phase using a small set of parallel multimodal instances, where semantically equivalent claim and evidence texts in the anchor and target languages are paired with the same visual evidence, extending cross-lingual alignment techniques to multimodal settings.

\paragraph{Per-sample shift estimation.} 
For each calibration instance $i$ with parallel claims $(c^i_{\ell_{\text{anc}}}, c^i_{\ell_c})$, parallel evidence $(e^i_{\ell_{\text{anc}}}, e^i_{\ell_e})$, and shared visual content $\mathrm{I}^i$, we extract hidden states at layer $l$ for $(a,b) \in \{(\ell_{\text{anc}},\ell_{\text{anc}}), (\ell_c,\ell_{\text{anc}}), (\ell_{\text{anc}},\ell_e)\}$:
\begin{equation}
\mathbf{h}^{(l),i}_{a,b} = f_\theta^{(l)}(c^i_a, e^i_b, \mathrm{I}^i).
\end{equation}

For each configuration, we compute mean representations over claim and evidence token spans $S_c$ and $S_e$\footnote{Spans are identified via explicit delimiter tokens in the prompt (see Appendix \ref{appendix:prompt}).}:
\begin{align}
\boldsymbol{\mu}^i_{c,a,b} &= \frac{1}{|S_c|} \sum_{t \in S_c} \mathbf{h}^{(l),i}_{a,b}[t], \\
\boldsymbol{\mu}^i_{e,a,b} &= \frac{1}{|S_e|} \sum_{t \in S_e} \mathbf{h}^{(l),i}_{a,b}[t].
\end{align}

To disentangle language-specific shifts, we hold one textual component fixed in $\ell_{\text{anc}}$ and measure the representation change when the other translates: claim from $\ell_{\text{anc}}$ to $\ell_c$, or evidence from $\ell_{\text{anc}}$ to $\ell_e$:
\begin{align}
\mathbf{v}^i_c &= \boldsymbol{\mu}^i_{c,\ell_{\text{anc}},\ell_{\text{anc}}} - \boldsymbol{\mu}^i_{c,\ell_c,\ell_{\text{anc}}}, \label{eq:shift_claim}\\
\mathbf{v}^i_e &= \boldsymbol{\mu}^i_{e,\ell_{\text{anc}},\ell_{\text{anc}}} - \boldsymbol{\mu}^i_{e,\ell_{\text{anc}},\ell_e}. \label{eq:shift_evidence}
\end{align}
Both shifts are defined as anchor minus target with $(\ell_{\text{anc}}, \ell_{\text{anc}})$ as the reference, so each vector points from the target language toward the anchor space and is applied additively in Equations~\ref{eq:steer_claim} and \ref{eq:steer_evidence}, as shown in Appendix~\ref{app:sign_ablation}.

\paragraph{Sign-aligned aggregation.} Following~\citet{zou2025representationengineeringtopdownapproach}, we prevent destructive cancellation by aligning vector signs before averaging  (analyzed in Appendix~\ref{app:component_ablation}). While the original RepE formulation uses binary task labels to determine sign direction, our calibration phase contains no task labels. We therefore adapt this step to a label-free setting by estimating a reference direction from the first $k$ calibration samples and aligning each vector to it using dot-product similarity. For each component $x \in \{c, e\}$:
\begin{enumerate}[leftmargin=*,nosep]
    \item Compute reference direction from the first $k$ samples:
    \begin{equation}
    \mathbf{v}_{x,\text{ref}} = \frac{1}{k} \sum_{i=1}^{k} \mathbf{v}^i_x.
    \end{equation}
    \item Determine alignment signs:
    \begin{equation}
    s^i_x = \operatorname{sign}\!\left( \langle \mathbf{v}^i_x, \mathbf{v}_{x,\text{ref}} \rangle \right).
    \end{equation}
    \item Aggregate with sign correction:
    \begin{equation}
    \mathbf{v}_x = \frac{1}{N} \sum_{i=1}^{N} s^i_x \cdot \mathbf{v}^i_x.
    \label{eq:sign_vector}
    \end{equation}
\end{enumerate}

\subsection{Applying Steering Vectors at Inference}
\label{sec:inference}

At test time, given an instance with claim in language $\ell_c$ and evidence in language $\ell_e$, we apply both precomputed steering vectors via forward hooks at layer $l$:
\begin{align}
\tilde{\mathbf{h}}^{(l)}[t] &= \mathbf{h}^{(l)}[t] + \lambda_c \cdot \mathbf{v}_c, \quad \forall t \in S_c, \label{eq:steer_claim}\\
\tilde{\mathbf{h}}^{(l)}[t] &= \mathbf{h}^{(l)}[t] + \lambda_e \cdot \mathbf{v}_e, \quad \forall t \in S_e, \label{eq:steer_evidence}
\end{align}
where the per-component strengths $\lambda_c, \lambda_e$ are derived automatically from the relative vector norms to avoid manual tuning. Let $n_c = \|\mathbf{v}_c\|$ and $n_e = \|\mathbf{v}_e\|$. For $n_c + n_e \ge 10^{-8}$, we set:
\begin{equation}
\begin{aligned}
\lambda_c &= \lambda \cdot \frac{n_e}{n_c + n_e}, \quad \lambda_e = \lambda \cdot \frac{n_c}{n_c + n_e}, && \text{if } n_c, n_e > 0, \\
\lambda_c & = 0, \quad \lambda_e = \lambda, && \text{if } n_c = 0, \\
\lambda_c & = \lambda, \quad \lambda_e = 0, &&  \text{if } n_e = 0,
\end{aligned}
\label{eq:adaptive_lambda}
\end{equation}
with a uniform fallback $\lambda_c = \lambda_e = \lambda/2$ when $n_c + n_e < 10^{-8}$. When both spans are non-anchor, this inverse-norm weighting ensures the smaller vector receives proportionally stronger scaling, preventing under-steering of weakly shifted components (see Appendix~\ref{app:component_ablation}). When only one span is non-anchor, the full strength $\lambda$ is applied to that span. 
Steering vectors are broadcast across batch and sequence dimensions; modified representations propagate through remaining layers without parameter updates.

\section{Experiments}
\label{sec:experiments}

\subsection{Setup}
\label{sec:setup}

\paragraph{Dataset.}
We evaluate on MOCHEG-XL, our machine-translated extension of MOCHEG~\citep{yao-etal-2023-mocheg} spanning nine languages. We use 2,442 test instances and split the 1,490 development instances into 372 calibration and 1,118 validation samples. Each instance contains a claim, textual evidence, visual evidence, and a three-way label: \textit{Supported}, \textit{Refuted}, or \textit{NEI (Not Enough Information)}.

\paragraph{Model and configuration.}
We evaluate Qwen3-VL 4B, Gemma 4 E2B, and LFM2-VL 3B. Steering vectors are estimated via sign-aligned aggregation in Section~\ref{sec:steering_vectors}, with key hyperparameters selected mainly from Qwen3-VL validation experiments, while LFM2-VL and Gemma 4 are evaluated to demonstrate generality. Full implementation and hyperparameter details are provided in Appendix~\ref{appendix:implementation_details}. 
We assess significance using paired bootstrap resampling over test instances with 10{,}000 resamples, matching the aggregation level of the reported tier means.

\paragraph{Baselines.}
We compare ORBIT with CLAIM~\citep{ye-etal-2025-claim} and two translation-based frameworks. 
\textbf{CLAIM}\footnote{Since no official public code is available, we re-implement it following their paper configurations.} trains per-head linear SVM probes on visual-masked attention outputs to select the top-$k$ heads and adds English-to-target shift vectors at the final input token during inference.
\textbf{Self-translate CoT} (Chain-of-Thought) prompts the base model to translate non-English text into English before verification. 
\textbf{Pivot NLLB} translates all input text using NLLB-200 distilled 600M\footnote{\href{https://huggingface.co/facebook/nllb-200-distilled-600M}{facebook/nllb-200-distilled-600M}}~\citep{costa2022nllb} before SVLM prediction. 
This distilled variant fits our resource-constrained scope, whereas larger or commercial translators form out-of-scope upper bounds. In addition, Google Cloud Translation covers only offline data and calibration, avoiding inference-time round-trip translation.

\subsection{Main Results}
\label{sec:results}

\begin{table*}[!t]
\centering
\small
\setlength{\tabcolsep}{3pt}
\renewcommand{\arraystretch}{1.08}
\resizebox{\textwidth}{!}{%
\begin{tabular}{lcccccc@{\hspace{0.25cm}}ccccc}
\toprule
\textbf{Method} 
& \multicolumn{6}{c}{\textbf{High-resource evidence}}
& \multicolumn{5}{c}{\textbf{Low-resource evidence}} \\
\cmidrule(lr){2-7} \cmidrule(l){8-12}
& en & es & ru & vi & ja & mean
& jw & cy & mi & ti & mean \\
\midrule

\multicolumn{12}{c}{\textbf{Qwen3-VL 4B}} \\ \hline

\multicolumn{12}{@{}l}{\textit{$\Delta$Accuracy}} \\
Pivot
& \msd{-1.48}{2.40} & \msd{-1.99}{1.88} & \msd{-0.50}{2.03} & \msd{-1.52}{2.01} & \msd{+0.61}{1.78} & \lmc{-0.98}
& \msd{+0.99}{1.34} & \msd{-1.28}{1.89} & \msd{+0.42}{1.74} & \bmsd{+2.64}{2.54} & \hmc{+0.69} \\

CoT
& \msd{-1.77}{2.03} & \msd{-0.34}{2.20} & \msd{+0.05}{1.61} & \msd{-0.08}{1.78} & \msd{+0.84}{1.66} & \lmc{-0.26}
& \msd{+1.91}{1.66} & \msd{+0.64}{2.18} & \msd{+0.62}{1.24} & \msd{+0.20}{1.56} & \hmc{+0.84} \\

CLAIM
& \msd{+0.01}{1.36} & \msd{-0.03}{0.78} & \msd{+0.05}{0.99} & \msd{+0.55}{0.45} & \msd{+0.92}{0.82} & \hmc{+0.30}
& \msd{+0.76}{2.79} & \msd{+0.27}{2.34} & \bmsd{+1.84}{0.99} & \msd{+0.89}{1.13} & \hmc{+0.94} \\

\textbf{ORBIT}
& \bmsd{+0.49}{1.21} & \bmsd{+0.80}{0.82} & \bmsd{+1.44}{0.92} & \bmsd{+1.37}{0.62} & \bmsd{+2.09}{1.22} & \hmcOne{\textbf{+1.24}}
& \bmsd{+2.70}{1.27} & \bmsd{+2.22}{1.12} & \msd{+1.80}{0.99} & \msd{+1.76}{0.96} & \hc{\textbf{+2.12}} \\

\midrule
\multicolumn{12}{@{}l}{\textit{$\Delta$F1-score}} \\
Pivot
& \msd{-3.08}{2.45} & \msd{-3.98}{2.13} & \msd{-2.51}{2.38} & \msd{-3.28}{2.32} & \msd{-1.50}{2.70} & \lc{-2.87}
& \msd{-0.40}{1.53} & \msd{-2.13}{2.70} & \msd{-1.80}{2.27} & \bmsd{+3.05}{1.44} & \lmc{-0.32} \\

CoT
& \msd{-3.20}{3.81} & \msd{-2.02}{3.56} & \msd{-1.67}{2.65} & \msd{-1.77}{3.08} & \msd{-1.25}{2.20} & \lc{-1.98}
& \msd{+0.40}{2.50} & \msd{-0.37}{2.98} & \msd{+0.17}{2.37} & \msd{-2.96}{2.46} & \lmc{-0.69} \\

CLAIM
& \msd{-1.71}{1.67} & \msd{-1.39}{0.94} & \msd{-1.40}{0.81} & \msd{-0.75}{0.70} & \msd{-0.46}{0.49} & \lc{-1.14}
& \msd{+0.02}{2.87} & \msd{-0.12}{2.39} & \bmsd{+2.61}{1.40} & \msd{+1.77}{1.91} & \hmcOne{+1.07} \\

\textbf{ORBIT}
& \bmsd{-1.63}{2.01} & \bmsd{-1.09}{1.37} & \bmsd{-0.53}{1.15} & \bmsd{-0.44}{1.20} & \bmsd{+0.16}{1.42} & \lmc{\textbf{-0.71}}
& \bmsd{+1.70}{1.14} & \bmsd{+1.84}{0.73} & \msd{+2.56}{1.31} & \msd{+2.68}{1.21} & \hc{\textbf{+2.20}} \\


\midrule

\multicolumn{12}{c}{\textbf{Gemma 4 E2B}} \\ \hline

\multicolumn{12}{@{}l}{\textit{$\Delta$Accuracy}} \\
Pivot
& \msd{-1.32}{1.31} & \msd{+0.10}{1.63} & \msd{+1.05}{0.93} & \msd{+0.41}{1.45} & \msd{+1.44}{1.43} & \hmc{+0.34}
& \msd{+1.09}{1.09} & \msd{+0.80}{1.96} & \msd{-0.27}{1.90} & \bmsd{+3.23}{2.11} & \hmcOne{+1.21} \\

CoT
& \msd{-1.87}{1.56} & \msd{+0.77}{1.28} & \msd{+0.94}{1.11} & \msd{+0.72}{1.46} & \msd{+0.74}{1.33} & \hmc{+0.26}
& \msd{+0.41}{1.68} & \msd{+0.12}{1.67} & \msd{+0.06}{1.94} & \msd{+0.73}{1.94} & \hmc{+0.33} \\

CLAIM
& \msd{+0.08}{0.77} & \msd{+0.72}{0.71} & \msd{+0.28}{1.12} & \msd{+0.50}{0.88} & \msd{-1.24}{0.67} & \hmc{+0.07}
& \msd{+0.52}{1.09} & \msd{+1.15}{0.54} & \msd{+1.02}{0.55} & \msd{+0.49}{0.41} & \hmc{+0.79} \\

\textbf{ORBIT}
& \bmsd{+1.48}{0.90} & \bmsd{+2.59}{1.20} & \bmsd{+2.54}{0.63} & \bmsd{+2.61}{1.06} & \bmsd{+2.32}{0.63} & \hc{\textbf{+2.31}}
& \bmsd{+2.45}{0.80} & \bmsd{+1.45}{1.01} & \bmsd{+1.39}{1.08} & \msd{+1.34}{1.17} & \hmcOne{\textbf{+1.66}} \\

\midrule
\multicolumn{12}{@{}l}{\textit{$\Delta$F1-score}} \\
Pivot
& \msd{-1.16}{1.29} & \msd{-0.20}{1.47} & \bmsd{+1.07}{1.14} & \msd{+0.34}{1.42} & \bmsd{+1.42}{2.10} & \hmc{+0.37}
& \msd{+1.77}{1.44} & \msd{+0.61}{2.39} & \msd{-2.32}{2.59} & \bmsd{+4.61}{3.43} & \hmcOne{+1.17} \\

CoT
& \msd{-1.81}{1.65} & \msd{+0.43}{1.12} & \msd{+0.39}{0.92} & \msd{+0.26}{1.40} & \msd{+0.31}{1.19} & \lmc{-0.08}
& \msd{+1.00}{1.59} & \msd{+1.36}{2.19} & \msd{+1.27}{2.32} & \msd{+2.14}{3.07} & \hmcOne{+1.44} \\

CLAIM
& \msd{+0.09}{0.92} & \msd{+0.70}{0.82} & \msd{-0.05}{1.64} & \msd{+0.19}{1.33} & \msd{-1.62}{0.82} & \lmc{-0.14}
& \msd{+0.49}{1.67} & \msd{+0.36}{1.11} & \msd{+0.44}{1.22} & \msd{-0.86}{1.18} & \hmc{+0.11} \\

\textbf{ORBIT}
& \bmsd{+0.58}{1.15} & \bmsd{+1.30}{2.04} & \msd{+0.57}{1.37} & \bmsd{+0.65}{1.77} & \msd{+0.58}{1.44} & \hmc{\textbf{+0.74}}
& \bmsd{+2.86}{0.75} & \bmsd{+2.40}{1.25} & \bmsd{+2.26}{1.22} & \msd{+4.05}{0.81} & \hc{\textbf{+2.89}} \\

\midrule

\multicolumn{12}{c}{\textbf{LFM2-VL 3B}} \\ \hline

\multicolumn{12}{@{}l}{\textit{$\Delta$Accuracy}} \\
Pivot
& \msd{+0.65}{1.84} & \msd{+0.09}{1.89} & \bmsd{+2.78}{1.94} & \msd{+3.77}{1.66} & \msd{+1.40}{2.39} & \hmcOne{+1.74}
& \msd{+2.57}{1.14} & \msd{+1.13}{0.86} & \msd{+1.39}{1.06} & \bmsd{+5.56}{1.16} & \hc{+2.66} \\

CoT
& \msd{-2.31}{1.61} & \msd{-1.41}{2.09} & \msd{+1.35}{1.62} & \msd{+3.53}{2.06} & \msd{+1.47}{2.57} & \hmc{+0.53}
& \msd{+2.98}{1.48} & \msd{+2.18}{0.78} & \msd{+2.52}{1.01} & \msd{+2.13}{1.37} & \hc{+2.45} \\

CLAIM
& \msd{-2.03}{0.81} & \msd{-2.10}{1.10} & \msd{+0.23}{0.95} & \msd{+4.70}{1.08} & \msd{+0.19}{0.84} & \hmc{+0.20}
& \msd{+1.20}{1.35} & \msd{+2.50}{1.38} & \msd{+3.15}{1.68} & \msd{+4.31}{2.10} & \hc{+2.79} \\

\textbf{ORBIT}
& \bmsd{+2.11}{0.82} & \bmsd{+2.28}{1.16} & \msd{+2.65}{0.74} & \bmsd{+5.15}{1.31} & \bmsd{+2.78}{0.76} & \hmcThree{\textbf{+3.00}}
& \bmsd{+3.06}{1.34} & \bmsd{+4.42}{1.92} & \bmsd{+3.64}{1.18} & \msd{+4.30}{1.01} & \hmcThree{\textbf{+3.85}} \\

\midrule
\multicolumn{12}{@{}l}{\textit{$\Delta$F1-score}} \\
Pivot
& \msd{+0.85}{2.65} & \msd{+0.34}{2.65} & \msd{+3.12}{2.97} & \msd{+4.92}{2.08} & \msd{+1.24}{3.31} & \hc{+2.09}
& \msd{+3.75}{2.07} & \msd{+0.67}{1.62} & \msd{-0.33}{2.00} & \msd{+3.96}{2.15} & \hc{+2.01} \\

CoT
& \msd{-4.16}{2.97} & \msd{-2.37}{3.85} & \msd{+0.21}{3.23} & \msd{+5.23}{3.37} & \msd{+3.23}{4.21} & \hmc{+0.43}
& \msd{+3.39}{3.16} & \msd{+1.27}{1.68} & \msd{+1.40}{1.96} & \msd{-2.32}{1.30} & \hmc{+0.93} \\

CLAIM
& \msd{-4.69}{1.67} & \msd{-4.26}{1.66} & \msd{-1.29}{1.43} & \msd{+3.17}{1.09} & \msd{-1.04}{1.28} & \lc{-1.62}
& \msd{+1.13}{1.34} & \msd{+3.78}{1.17} & \msd{+2.81}{1.83} & \bmsd{+4.31}{1.74} & \hmcThree{+3.01} \\

\textbf{ORBIT}
& \bmsd{+2.98}{1.09} & \bmsd{+3.40}{1.21} & \bmsd{+3.64}{0.99} & \bmsd{+5.69}{1.68} & \bmsd{+3.66}{1.03} & \hmcThree{\textbf{+3.87}}
& \bmsd{+4.33}{2.18} & \bmsd{+5.30}{2.47} & \bmsd{+4.58}{1.74} & \msd{+4.01}{0.61} & \hmcThree{\textbf{+4.55}} \\

\bottomrule
\end{tabular}
}
\caption{Comparison of ORBIT with CLAIM~\citep{ye-etal-2025-claim} and translation-based baselines across evidence languages. Pivot refers to Pivot NLLB and CoT refers to Self-translate CoT. Values report mean $\pm$ standard deviation for $\Delta$Accuracy and $\Delta$F1-score relative to the zero-shot baseline, averaged over claim languages. Bold indicates the best result per language and metric. Full per-configuration results are provided in Appendix~\ref{appendix:full_results}.}
\label{tab:main_results}
\end{table*}

Table~\ref{tab:main_results} reports changes in accuracy and macro F1-score relative to the zero-shot baseline. For each evidence language, we average results over all claim languages; the standard deviation (SD) reflects how sensitive each method is to claim-language variation. We prioritize per-evidence aggregations, as they directly isolate the source of cross-lingual degradation; the complete 81-configuration matrix is provided in Appendix~\ref{appendix:full_results}.

\paragraph{Low-resource recovery.} 
ORBIT substantially recovers low-resource evidence, pulling collapsed representations toward the English anchor. Gains are model-dependent: LFM2-VL benefits most across both tiers, Gemma 4 improves steadily, while Qwen3-VL shows more modest recovery and incurs a high-resource F1 drop. 
For Qwen3-VL, paired bootstrap resampling confirms these low-resource gains remain statistically significant despite a smaller magnitude (Appendix~\ref{appendix:bootstrap_low_resource}), while the model's high-resource F1 degradation is driven primarily by the \textit{NEI} class (Appendix~\ref{appendix:classification_report}). This occurs because overshooting erases the representational ambiguity required for abstention and pushes borderline instances toward \textit{Supported} or \textit{Refuted}. Control experiments with mismatched-language vectors and label-bias calibration in Appendix~\ref{app:control_experiments} confirm that ORBIT's gains stem from language-matched steering rather than a mere label-prior shift.

\paragraph{Evidence-language asymmetry.} The gains follow the same asymmetry observed in the zero-shot collapse: evidence language, rather than claim language, is the dominant source of performance variation. Evidence passages are longer and carry the factual content needed to ground the image, whereas claims act as compact verification hypotheses. Consequently, ORBIT yields consistent accuracy improvements across claim languages (full results in Appendix~\ref{appendix:full_results}), while the largest recoveries occur when the evidence language is low-resource. The claim language itself matters less than the evidence language.

\paragraph{Comparison with baselines.}
ORBIT outperforms CLAIM, Self-translate CoT, and Pivot NLLB across three evaluated model families. To be specific, CLAIM's probe-selected, final-token attention intervention and uniform per-language scaling struggle on long textual documents. By contrast, ORBIT is probe-free, steers full claim--evidence hidden spans, and balances mismatched pairs via inverse-norm weighting.
Despite differing in whether translation is handled internally or by an external model, both translation baselines share a translate-then-verify paradigm that has inconsistent gains across languages. Translation noise on long passages disrupts cross-modal grounding and incurs a severe latency overhead. By operating directly on hidden representations, ORBIT bypasses auxiliary decoding to deliver more robust grounding recovery. However, the impact of static per-language steering is model-dependent. Although LFM2-VL and Gemma 4 achieve consistent gains across both resource tiers, uniform vectors may over-steer already well-aligned high-resource representations in Qwen3-VL. This contrast exposes a clear trade-off between low-resource recovery and high-resource preservation.

\subsection{Calibration Norms as a Diagnostic Proxy}
\label{sec:norm_analysis}

\begin{figure}[t]
    \centering
    \begin{subfigure}[b]{0.48\textwidth}
        \centering
        \includegraphics[width=\linewidth]{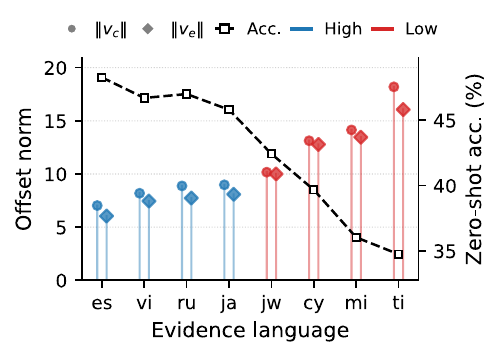}
        \caption{}
        \label{fig:norm_diagnostic}
    \end{subfigure}
    \hfill
    \begin{subfigure}[b]{0.48\textwidth}
        \centering
        \includegraphics[width=\linewidth]{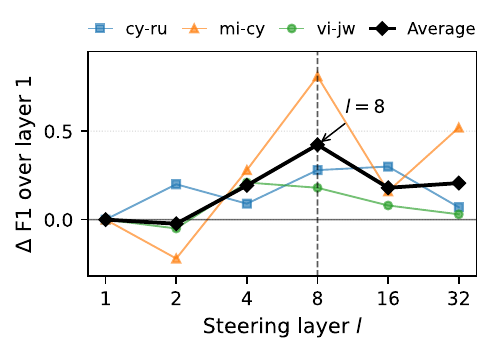}
        \caption{}
        \label{fig:layer_selection}
    \end{subfigure}
    \caption{Diagnostic analyses on Qwen3-VL: (a) Offset norms and zero-shot accuracy by evidence language. Accuracy is averaged over all claim languages for each fixed evidence language. (b) Layer-selection ablation on the 1,118-instance validation split. $\Delta$F1 is measured relative to layer~1.}
    \label{fig:combined_diagnostic}
\end{figure}

Figure~\ref{fig:norm_diagnostic} shows that evidence offset norms from 372 calibration samples strongly predict zero-shot accuracy collapse in Figure~\ref{fig:cross_lingual_heatmap}, yielding a Pearson correlation of $-0.986$ and enabling lightweight triage without full-test evaluation. Claim norms exceed evidence norms, yet evidence-side steering drives recovery because evidence carries the factual content required for visual grounding.
This strong correlation holds because offset norms directly quantify the representational gap between anchor and target language hidden states prior to inference. 
Appendix~\ref{app:calibration_sensitivity} reports a representative calibration-size sensitivity check on Qwen3-VL with Vietnamese claims and low-resource evidence languages.

\subsection{Layer Selection}
\label{sec:layer_selection}

We evaluate steering layers $l \in \{1,2,4,8,16,32\}$ on the validation split for Qwen3-VL (Figure~\ref{fig:layer_selection}). Performance peaks at $l=8$, but ORBIT remains stable across layers, with mean F1 varying by less than 0.5 points. This suggests that the method does not rely on a finely tuned intervention depth. Mid-layer steering is slightly stronger, likely because multilingual representations are already partially composed but not yet fully specialized toward the final verdict. Sensitivity is higher for severely misaligned pairs such as mi--cy ($+0.81$ from layer 1 to 8), while vi--jw changes only modestly ($+0.18$).

\subsection{Efficiency Analysis}

Table~\ref{tab:efficiency} compares inference time of Qwen3-VL across methods on the test set. Self-translate CoT is prohibitively slow, taking nearly 20$\times$ longer than zero-shot because it must autoregressively translate long evidence passages before verification. Pivot NLLB is faster but still requires a separate translation model, increasing memory use and pipeline complexity.
CLAIM is similarly lightweight at inference, but it requires a calibration phase on the same 372 calibration samples as ORBIT, in which it probes attention heads and estimates language shift vectors from paired English and target-language inputs, and it yields a smaller low-resource gain. ORBIT avoids external components and applies steering through a lightweight forward hook, adding 19\% overhead over zero-shot while achieving the largest low-resource accuracy gain. This makes ORBIT the fastest non-baseline intervention and more practical for resource-constrained deployment.

\begin{table}
\centering
\small
\begin{tabular}{@{}lcc@{}}
\toprule
\textbf{Method} & \textbf{Inference Time} & \textbf{$\Delta$Accuracy\hspace{0.05cm}\textsubscript{low-res}} \\
\midrule
Zero-shot baseline & \textbf{1057.28s} & -- \\
Pivot NLLB & 2007.45s & +0.69 \\
Self-translate CoT & 20888.31s & \textit{+0.84} \\

CLAIM & 1290.44s & +0.94 \\
\midrule
ORBIT (Ours) & \textit{1258.62s} & \textbf{+2.12} \\
\bottomrule
\end{tabular}
\caption{Mean inference time on the test set using Qwen3-VL. ORBIT introduces moderate overhead versus the zero-shot baseline but remains significantly faster and more resource-efficient than other baselines.}
\label{tab:efficiency}
\end{table}

\section{Conclusion}
\label{sec:conclusion}

We identify and quantify a severe, resource-asymmetric collapse in small vision-language models for cross-lingual multimodal fact verification. While high-resource languages degrade modestly, low-resource evidence languages drive accuracy to near-random levels despite intact visual encoding. To address this without gradient updates, we propose ORBIT, a training-free activation steering method that realigns target-language representations toward the model's high-resource anchor space using per-language correction vectors estimated from parallel bilingual pairs. ORBIT partially recovers low-resource performance while preserving capability on high-resource languages, establishing that inference-time representation adjustment can bridge cross-lingual gaps when training resources are limited.

\section*{Limitations}
\label{sec:limitations}

ORBIT exhibits a model-dependent trade-off. Static per-language steering vectors applied uniformly cannot simultaneously preserve well-aligned high-resource representations and correct severely misaligned low-resource ones. As shown in Section~\ref{sec:results}, this trade-off primarily affects Qwen3-VL, where high-resource evidence incurs F1 penalties -- especially on the NEI class. In contrast, low-resource evidence benefits from realignment across all models. Future work should explore instance-specific modulation based on estimated representation quality rather than coarse resource-tier heuristics, though reliably estimating such quality at inference time without ground-truth supervision remains non-trivial.

Our evaluation is intentionally controlled: MOCHEG-XL is machine-translated and serves as a diagnostic instrument rather than a human-verified benchmark. Consequently, it does not capture noisier evidence, code-switching, culturally specific visuals, or multilingual multi-source settings found in real-world fact-checking. We pair each claim with single-language evidence and do not yet aggregate conflicting cross-lingual signals. 
Steering vectors estimated from machine-translated instances inherit translation noise, especially for severely low-resource languages. This warrants investigation into higher-quality calibration sources and robustness tests under systematic noise. 
Ultimately, while Appendix~\ref{appendix:translation_quality} and ORBIT's gains on identical translated inputs suggest that the collapse is not merely translation noise, holding visual evidence constant cannot fully separate cross-modal grounding from text comprehension errors.

\subsection*{AI use statement}

In this work, we used ChatGPT 5.5, Claude Sonnet 4.5, and Gemini 3.1 Pro for language polishing, grammar checking, sentence-level clarity improvements, and coding assistance. We have reviewed all AI-assisted work. We take responsibility for the final content of this work, including text, claims or artifacts produced with the aid of generative AI.




\subsection*{Ethics statement}

This work addresses cross-lingual inequity in multimodal fact verification using publicly available artifacts under their respective licenses. MOCHEG-XL is a machine-translated diagnostic extension designed for controlled analysis of linguistic shift, not a human-verified benchmark. Upon acceptance, we will release MOCHEG-XL and ORBIT implementation under an open license to support reproducible, equitable multilingual research.

\bibliography{iclr2027_conference}
\bibliographystyle{iclr2027_conference}

\clearpage

\appendix

\section*{Appendices}
\label{sec:appendix}

\section{MOCHEG statistics}
\label{appendix:mocheg_stats}

\begin{table}[H]
\centering
\small
\renewcommand{\arraystretch}{1.15}
\begin{tabular}{lrr}
\toprule
\textbf{Statistic} & \textbf{Dev} & \textbf{Test} \\
\midrule

Number of claims & 1,490 & 2,442 \\
\midrule

\multicolumn{3}{@{}l}{\textit{\textbf{Evidence Distribution}}} \\
\hspace{1em}Number of text paragraphs & 4,067 & 6,268 \\
\hspace{1em}Number of images & 1,178 & 2,007 \\
\midrule

\multicolumn{3}{@{}l}{\textit{\textbf{Label Distribution}}} \\
\hspace{1em}Refuted & 488 & 825 \\
\hspace{1em}Supported & 501 & 817 \\
\hspace{1em}NEI & 501 & 800 \\

\bottomrule
\end{tabular}
\caption{Detailed statistics of the MOCHEG dataset splits used in our experiments.}
\label{tab:dataset_stats}
\end{table}

\begin{figure}[H]
    \centerline{\includegraphics[width=1\textwidth]{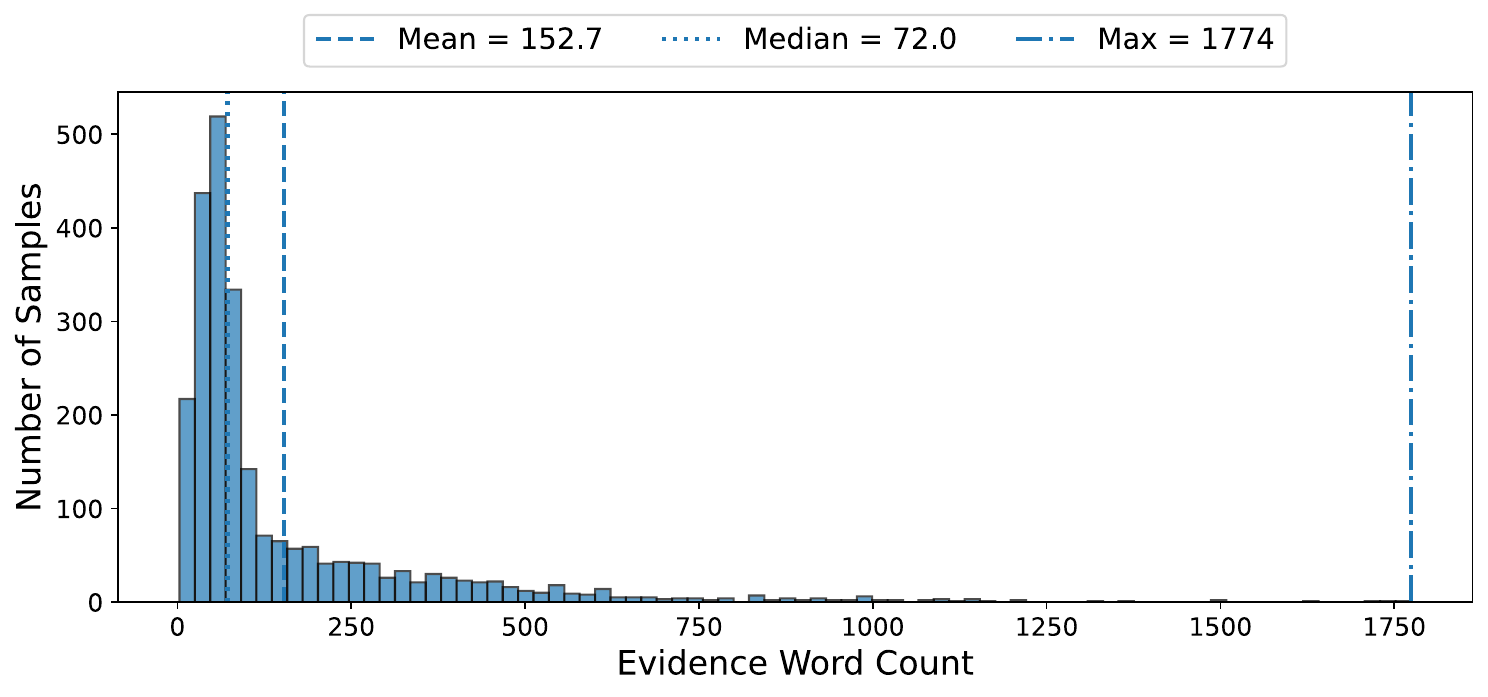}}
    \caption{Word count distribution of textual evidence in the MOCHEG testing set. URLs are counted as single tokens, and words with hyphens or apostrophes are preserved as single units.}
    \label{fig:mocheg_length_dist}
\end{figure}

\section{Translation Quality}
\label{appendix:translation_quality}

We evaluate translation quality using GPT-5.4 Mini~\citep{openai2026gpt54} and Gemini 3.1 Flash-Lite~\citep{gemini3.1flashlite} as LLM judges~\citep{NEURIPS2023_91f18a12}.

\subsection{Prompt}
\label{appendix:translation_quality_prompt}

\begin{lstlisting}[style=terminal]
Evaluate whether the translated text preserves the meaning of the English text.

Consider the overall meaning and important factual details. Allow natural paraphrasing, reordering, and language-specific phrasing. Mark as not adequate if the translation omits important information, adds unsupported information, changes facts, reverses meaning, or is too incomplete/unclear to verify.

Return ONLY valid JSON:
{{
  "adequate": true/false,
  "score": 0-100,
  "reason": "brief explanation",
  "major_issues": []
}}

Use "major_issues" only for important meaning changes. Leave it empty if the translation is adequate.

English text:
{english_text}

Translated text:
{translated_text}
\end{lstlisting}

\subsection{Results}



\begin{table}[H]
\centering
\resizebox{\linewidth}{!}{%
\setlength{\tabcolsep}{4.5pt}
\begin{tabular}{@{}ll cccc cccc@{}}
\toprule
\multirow{2}{*}{\textbf{Model}} & \multirow{2}{*}{\textbf{Metric}} & \multicolumn{4}{c}{\textbf{High-Resource}} & \multicolumn{4}{c}{\textbf{Low-Resource}} \\
\cmidrule(lr){3-6} \cmidrule(lr){7-10}
& & \textbf{es} & \textbf{ru} & \textbf{vi} & \textbf{ja} & \textbf{jw} & \textbf{cy} & \textbf{mi} & \textbf{ti} \\
\midrule
\multirow{2}{*}{GPT 5.4 Mini}          & Claim    & 96.86 & 94.42 & 95.72 & 91.55 & 89.49 & 95.59 & 87.61 & 92.06 \\
                                       & Evidence & 96.56 & 93.44 & 93.07 & 90.33 & 83.81 & 83.00 & 76.58 & 76.08 \\
\midrule
\multirow{2}{*}{Gemini 3.1 Flash-Lite} & Claim    & 98.10 & 93.40 & 95.31 & 93.53 & 86.75 & 95.50 & 86.10 & 93.62 \\
                                       & Evidence & 98.38 & 92.89 & 94.21 & 93.33 & 81.50 & 87.61 & 75.26 & 81.92 \\
\bottomrule
\end{tabular}%
}
\caption{Average LLM-as-a-judge evaluation scores on MOCHEG-XL across high-resource and low-resource languages.}
\label{tab:evaluation_scores}
\end{table}

Our evaluation protocol mitigates potential LLM-as-a-judge bias in low-resource languages by explicitly prioritizing factual alignment over syntactic fluency. Instead of evaluating the low-resource output in isolation, the prompt (Appendix~\ref{appendix:translation_quality_prompt}) uses the original English text as a reference to verify core entities and key facts. Furthermore, two distinct LLM judges produce highly consistent scores across all target languages (Table~\ref{tab:evaluation_scores}). This agreement indicates that the evaluations reflect actual semantic preservation rather than shared hallucinations. Consequently, while low-resource translation may introduce minor grammatical noise, the core facts remain intact. This ensures MOCHEG-XL remains a valid diagnostic instrument for isolating cross-modal grounding failures, rather than merely reflecting translation errors.

\section{Base Results}
\label{appendix:additional_base_results}

\begin{figure}[H]
\centering
\tiny
\setlength{\tabcolsep}{2.75pt}
\setlength{\fboxsep}{1pt}

\begin{minipage}[t]{0.5\textwidth}
\centering
\renewcommand{\arraystretch}{1.15}
\begin{tabular}{l|ccccccccc}
\toprule
\multirow{2}{*}{\textbf{Claim}} & \multicolumn{9}{c}{\textbf{Evidence Language}} \\
\cmidrule{2-10}
& \textbf{en} & \textbf{es} & \textbf{ru} & \textbf{vi} & \textbf{ja} & \textbf{jw} & \textbf{cy} & \textbf{mi} & \textbf{ti} \\
\midrule
\textbf{en} & \marked[acc-tl]{\hc{48.08}} & \hc{46.60} & \hc{45.17} & \hc{45.21} & \marked[acc-tr]{\hc{45.99}} & \mc{39.72} & \mc{37.10} & \mc{38.04} & \mc{35.34} \\
\textbf{es} & \hc{49.63} & \hc{48.03} & \hc{45.66} & \hc{45.45} & \mc{44.92} & \mc{39.68} & \mc{36.32} & \mc{36.77} & \lc{34.81} \\
\textbf{ru} & \hc{48.24} & \hc{45.78} & \hc{46.03} & \hc{45.13} & \mc{44.76} & \mc{40.25} & \mc{35.30} & \mc{36.12} & \lc{34.03} \\
\textbf{vi} & \hc{47.58} & \hc{45.13} & \mc{44.88} & \hc{46.72} & \hc{45.29} & \mc{39.68} & \mc{35.01} & \mc{36.40} & \lc{34.40} \\
\textbf{ja} & \marked[acc-bl]{\hc{48.24}} & \hc{46.15} & \mc{44.27} & \mc{44.39} & \marked[acc-br]{\hc{46.89}} & \mc{38.78} & \lc{34.11} & \mc{35.22} & \lc{33.09} \\
\textbf{jw} & \mc{44.14} & \mc{41.65} & \mc{41.89} & \mc{40.91} & \mc{41.85} & \mc{41.44} & \lc{33.42} & \mc{35.67} & \lc{32.88} \\
\textbf{cy} & \mc{41.69} & \mc{37.96} & \mc{37.96} & \mc{37.22} & \mc{38.21} & \mc{36.40} & \mc{38.12} & \lc{34.73} & \lc{32.96} \\
\textbf{mi} & \mc{43.00} & \mc{39.23} & \mc{38.86} & \mc{39.23} & \mc{39.89} & \mc{36.04} & \lc{33.78} & \mc{39.35} & \lc{33.58} \\
\textbf{ti} & \mc{43.24} & \mc{38.86} & \mc{38.57} & \mc{39.35} & \mc{38.78} & \mc{36.61} & \lc{33.70} & \lc{34.44} & \mc{36.86} \\
\bottomrule
\end{tabular}
\caption*{Accuracy (\%)}
\end{minipage}%
\hfill
%
\begin{minipage}[t]{0.5\textwidth}
\centering
\renewcommand{\arraystretch}{1.15}
\begin{tabular}{l|ccccccccc}
\toprule
\multirow{2}{*}{\textbf{Claim}} & \multicolumn{9}{c}{\textbf{Evidence Language}} \\
\cmidrule{2-10}
& \textbf{en} & \textbf{es} & \textbf{ru} & \textbf{vi} & \textbf{ja} & \textbf{jw} & \textbf{cy} & \textbf{mi} & \textbf{ti} \\
\midrule
\textbf{en} & \marked[f1-tl]{\hc{46.43}} & \hc{45.18} & \mc{44.25} & \mc{44.27} & \marked[f1-tr]{\hc{45.41}} & \mc{38.52} & \mc{35.82} & \mc{36.94} & \lc{33.02} \\
\textbf{es} & \hc{48.34} & \hc{46.25} & \hc{45.54} & \hc{45.67} & \hc{45.03} & \mc{39.15} & \mc{35.27} & \mc{35.67} & \lc{32.03} \\
\textbf{ru} & \hc{46.97} & \hc{45.29} & \mc{44.92} & \hc{45.23} & \mc{44.89} & \mc{39.60} & \lc{33.81} & \lc{34.80} & \lc{31.22} \\
\textbf{vi} & \hc{46.45} & \mc{44.86} & \hc{45.12} & \hc{45.90} & \hc{45.46} & \mc{38.76} & \lc{33.15} & \mc{35.06} & \lc{30.80} \\
\textbf{ja} & \marked[f1-bl]{\hc{46.94}} & \hc{45.78} & \mc{44.31} & \mc{44.50} & \marked[f1-br]{\hc{46.64}} & \mc{38.03} & \lc{32.28} & \lc{33.71} & \lc{29.59} \\
\textbf{jw} & \mc{43.97} & \mc{41.88} & \mc{42.17} & \mc{41.34} & \mc{42.22} & \mc{40.71} & \lc{31.52} & \lc{34.27} & \lc{29.62} \\
\textbf{cy} & \mc{41.93} & \mc{38.19} & \mc{38.04} & \mc{37.35} & \mc{38.24} & \mc{35.54} & \mc{37.63} & \lc{32.97} & \lc{29.79} \\
\textbf{mi} & \mc{43.23} & \mc{39.63} & \mc{39.26} & \mc{39.62} & \mc{40.18} & \mc{35.18} & \lc{32.07} & \mc{39.09} & \lc{30.67} \\
\textbf{ti} & \mc{43.54} & \mc{39.19} & \mc{38.60} & \mc{39.33} & \mc{38.68} & \mc{35.11} & \lc{30.92} & \lc{31.65} & \mc{35.52} \\
\bottomrule
\end{tabular}
\caption*{F1-score (\%)}
\end{minipage}

\begin{tikzpicture}[remember picture, overlay]
  \draw[black, line width=0.45pt] 
    ($(acc-bl)+(-0.34,-0.155)$) -- ($(acc-br)+(0.34,-0.155)$) -- ($(acc-tr)+(0.34,0.1275)$);
    
  \draw[black, line width=0.45pt] 
    ($(f1-bl)+(-0.34,-0.155)$) -- ($(f1-br)+(0.34,-0.155)$) -- ($(f1-tr)+(0.34,0.1275)$);
\end{tikzpicture}

\vspace{0.3cm}

\begin{minipage}{\textwidth}
\centering
\scriptsize
\textbf{Performance levels:} 
\colorbox{highperf!30}{\strut~High ($>$\,45\%)~} 
\colorbox{medperf!40}{\strut~Medium (35--45\%)~} 
\colorbox{lowperf!50}{\strut~Low ($<$\,35\%)~}
\end{minipage}

\caption{Cross-lingual performance of Gemma 4 across 81 claim--evidence configurations.}
\label{fig:cross_lingual_gemma4_heatmap}
\end{figure}

\begin{figure}[H]
\centering
\tiny
\setlength{\tabcolsep}{2.75pt}
\setlength{\fboxsep}{1pt}

\begin{minipage}[t]{0.5\textwidth}
\centering
\renewcommand{\arraystretch}{1.15}
\begin{tabular}{l|ccccccccc}
\toprule
\multirow{2}{*}{\textbf{Claim}} & \multicolumn{9}{c}{\textbf{Evidence Language}} \\
\cmidrule{2-10}
& \textbf{en} & \textbf{es} & \textbf{ru} & \textbf{vi} & \textbf{ja} & \textbf{jw} & \textbf{cy} & \textbf{mi} & \textbf{ti} \\
\midrule
\textbf{en} & \marked[acc-tl]{\hc{46.81}} & \hc{44.88} & \mc{39.72} & \mc{34.64} & \marked[acc-tr]{\hc{40.66}} & \mc{32.43} & \mc{31.90} & \mc{30.96} & \mc{30.71} \\
\textbf{es} & \hc{45.54} & \hc{46.56} & \mc{39.35} & \mc{35.54} & \hc{41.36} & \mc{33.29} & \mc{32.47} & \mc{31.61} & \mc{30.79} \\
\textbf{ru} & \hc{43.04} & \hc{42.26} & \hc{41.89} & \mc{34.32} & \mc{39.93} & \mc{32.39} & \mc{30.88} & \mc{30.88} & \mc{30.10} \\
\textbf{vi} & \hc{41.56} & \mc{38.82} & \mc{35.71} & \mc{36.49} & \mc{38.17} & \mc{31.24} & \mc{30.47} & \mc{30.18} & \lc{29.44} \\
\textbf{ja} & \marked[acc-bl]{\hc{44.92}} & \hc{44.02} & \hc{40.09} & \mc{36.00} & \marked[acc-br]{\hc{44.80}} & \mc{32.60} & \mc{31.74} & \mc{31.20} & \mc{31.00} \\
\textbf{jw} & \mc{39.60} & \mc{39.56} & \mc{35.09} & \mc{33.91} & \mc{36.73} & \mc{31.45} & \mc{30.26} & \lc{29.77} & \lc{28.58} \\
\textbf{cy} & \mc{38.66} & \mc{37.02} & \mc{34.44} & \mc{32.96} & \mc{35.95} & \mc{31.33} & \mc{30.34} & \mc{30.84} & \lc{27.93} \\
\textbf{mi} & \mc{39.19} & \mc{37.80} & \mc{34.11} & \mc{32.39} & \mc{36.65} & \mc{31.70} & \mc{31.16} & \mc{32.19} & \lc{27.85} \\
\textbf{ti} & \mc{37.31} & \mc{36.53} & \mc{32.72} & \mc{32.23} & \mc{34.56} & \mc{32.15} & \mc{30.92} & \mc{30.22} & \lc{27.15} \\
\bottomrule
\end{tabular}
\caption*{Accuracy (\%)}
\end{minipage}%
\hfill
%
\begin{minipage}[t]{0.5\textwidth}
\centering
\renewcommand{\arraystretch}{1.15}
\begin{tabular}{l|ccccccccc}
\toprule
\multirow{2}{*}{\textbf{Claim}} & \multicolumn{9}{c}{\textbf{Evidence Language}} \\
\cmidrule{2-10}
& \textbf{en} & \textbf{es} & \textbf{ru} & \textbf{vi} & \textbf{ja} & \textbf{jw} & \textbf{cy} & \textbf{mi} & \textbf{ti} \\
\midrule
\textbf{en} & \marked[f1-tl]{\hc{45.26}} & \hc{42.93} & \mc{37.77} & \mc{31.29} & \marked[f1-tr]{\mc{38.41}} & \lc{28.64} & \lc{27.59} & \lc{26.56} & \lc{27.18} \\
\textbf{es} & \hc{43.63} & \hc{44.74} & \mc{37.30} & \mc{32.51} & \mc{38.67} & \mc{30.18} & \lc{28.22} & \lc{27.54} & \lc{27.84} \\
\textbf{ru} & \hc{41.06} & \mc{39.98} & \hc{40.62} & \mc{30.82} & \mc{37.48} & \lc{28.72} & \lc{26.58} & \lc{26.53} & \lc{26.66} \\
\textbf{vi} & \mc{38.62} & \mc{35.46} & \mc{31.72} & \mc{33.61} & \mc{34.24} & \lc{26.27} & \lc{24.57} & \lc{24.51} & \lc{25.04} \\
\textbf{ja} & \marked[f1-bl]{\hc{43.96}} & \hc{42.78} & \mc{38.51} & \mc{32.93} & \marked[f1-br]{\hc{43.69}} & \lc{29.51} & \lc{27.76} & \lc{27.14} & \lc{27.14} \\
\textbf{jw} & \mc{35.77} & \mc{35.50} & \mc{30.89} & \lc{29.22} & \mc{32.17} & \lc{28.08} & \lc{25.13} & \lc{24.78} & \lc{24.77} \\
\textbf{cy} & \mc{35.01} & \mc{33.11} & \lc{29.96} & \lc{27.59} & \mc{31.26} & \lc{25.93} & \lc{26.20} & \lc{25.52} & \lc{24.04} \\
\textbf{mi} & \mc{35.42} & \mc{33.47} & \lc{29.40} & \lc{27.02} & \mc{31.61} & \lc{26.75} & \lc{25.61} & \lc{28.16} & \lc{23.84} \\
\textbf{ti} & \mc{33.10} & \mc{32.10} & \lc{27.19} & \lc{26.32} & \lc{29.48} & \lc{25.95} & \lc{23.63} & \lc{23.08} & \lc{21.95} \\
\bottomrule
\end{tabular}
\caption*{F1-score (\%)}
\end{minipage}

\begin{tikzpicture}[remember picture, overlay]
  \draw[black, line width=0.45pt] 
    ($(acc-bl)+(-0.34,-0.155)$) -- ($(acc-br)+(0.34,-0.155)$) -- ($(acc-tr)+(0.34,0.1275)$);
    
  \draw[black, line width=0.45pt] 
    ($(f1-bl)+(-0.34,-0.155)$) -- ($(f1-br)+(0.34,-0.155)$) -- ($(f1-tr)+(0.34,0.1275)$);
\end{tikzpicture}

\vspace{0.3cm}

\begin{minipage}{\textwidth}
\centering
\scriptsize
\textbf{Performance levels:} 
\colorbox{highperf!30}{\strut~High ($>$\,40\%)~} 
\colorbox{medperf!40}{\strut~Medium (30--40\%)~} 
\colorbox{lowperf!50}{\strut~Low ($<$\,30\%)~}
\end{minipage}

\caption{Cross-lingual performance of LFM2-VL across 81 claim--evidence configurations.}
\label{fig:cross_lingual_lfm2_heatmap}
\end{figure}

Corroborating the trends observed for Qwen3-VL in Section~\ref{sec:cross_lingual_collapse}, both Gemma 4 and LFM2-VL exhibit a severe evidence-language collapse (Figures \ref{fig:cross_lingual_gemma4_heatmap} and \ref{fig:cross_lingual_lfm2_heatmap}). When evidence shifts to low-resource languages (jw, cy, mi, ti), performance across all claim languages degrades to near-random levels. Crucially, the same-language diagonal advantage present in high-resource settings disappears entirely. These findings confirm that cross-modal grounding failure under low-resource conditions is a generalized vulnerability of SVLMs, rather than an isolated architectural artifact.

\section{Prediction Prompt}
\label{appendix:prompt}
\begin{lstlisting}[style=terminal]
Your task is to verify the truthfulness of the following claim using the provided text evidence and the accompanying image evidence.

IMPORTANT: The marker strings are for processing only; ignore them.

Classify the claim into exactly one of the following categories:
- Supported
- Refuted
- NEI (Not Enough Information)

Claim:
<CLAIM_START>
{claim_text}
<CLAIM_END>

Text Evidence:
<EVID_START>
{evidence_text}
<EVID_END>

Output only the label (Supported, Refuted, or NEI).
\end{lstlisting}

\section{Full Results}
\label{appendix:full_results}

Figures~\ref{fig:orbit_delta_heatmap}--\ref{fig:claim_lfm2_delta_heatmap} report the complete per-configuration results across all 81 claim--evidence language pairs. Each cell shows the change in accuracy or macro F1-score relative to the zero-shot baseline. These full matrices support the aggregate trends in Table~\ref{tab:main_results}: evidence language drives most performance variation, ORBIT produces the clearest recovery under low-resource evidence, and translation-based baselines are less stable across the full language grid.

\section{Implementation Details}
\label{appendix:implementation_details}

\subsection{ORBIT Hyperparameters and Inference Settings}

Steering vectors are estimated from calibration samples using sign-aligned aggregation. We use the first $k=16$ samples to compute the reference direction. Hyperparameters are selected on the Qwen3-VL validation split and applied uniformly across all models to demonstrate generality. Steering is applied at layer 8, selected from $\{1,2,4,8,16,32\}$. We apply a single steering strength $\lambda=0.05$ across all evidence languages, tuned on the validation split from $\{0.05,0.1,0.2,0.3,0.4,0.5\}$.

Images are resized so that the shortest side is 480 pixels while preserving the original aspect ratio. Generation uses greedy decoding with a maximum of 16 new tokens. All experiments are run on a single NVIDIA A100 SXM4 40GB GPU.

\subsection{Pivot NLLB Implementation}

For Pivot NLLB, we back-translate the non-English MOCHEG-XL claims and evidence into English, rather than using the original English MOCHEG inputs. Since our NLLB pipeline uses a 512-token maximum input length, long evidence passages are segmented with NLTK sentence tokenization before translation. Consecutive sentences are grouped into chunks of approximately 80 words, a conservative threshold chosen to keep low-resource language inputs below the token limit because words may split into multiple subword tokens.

Each chunk is translated independently with beam search using four beams, a repetition penalty of 1.15, and a no-repeat trigram constraint. Translated chunks are concatenated in their original order to reconstruct the English evidence passage. Empty evidence fields are left unchanged.

\subsection{Self-translate CoT Prompt}
\label{appendix:translation_prompt}

For the Self-translate CoT baseline, the base model is first prompted to translate non-English text into English before verification. We use the following translation prompt:

\begin{lstlisting}[style=terminal]
Translate the following text into English.
- Preserve the original meaning exactly.
- Keep names, dates, numbers, and units unchanged.
- Do NOT add explanations, notes, or any extra text.

Text:
{text}

Output ONLY the English translation.
\end{lstlisting}

\section{Bootstrap Significance Analysis}
\label{appendix:bootstrap_low_resource}

\begin{table}[H]
\centering
\footnotesize
\begin{tabular}{lcc}
\toprule
\textbf{Evidence} & \textbf{Accuracy 95\% CI} & \textbf{Macro F1 95\% CI} \\
\midrule
jw & [+1.82, +3.59] & [+0.89, +2.52] \\
cy & [+1.41, +3.02] & [+1.08, +2.61] \\
mi & [+0.98, +2.61] & [+1.78, +3.33] \\
ti & [+1.04, +2.46] & [+1.93, +3.43] \\
\midrule
\textbf{Tier mean} & [\textbf{+1.50}, \textbf{+2.73}] & [\textbf{+1.63}, \textbf{+2.78}] \\
\bottomrule
\end{tabular}
\caption{Paired bootstrap 95\% confidence intervals for ORBIT (Qwen3-VL) gains under low-resource evidence. All intervals exclude zero, indicating statistically significant improvements.}
\label{tab:bootstrap_low_resource}
\end{table}

Among the three evaluated models, Qwen3-VL exhibits the smallest low-resource gains (Table~\ref{tab:main_results}). Because it also serves as the primary backbone for our in-depth diagnostic and ablation analyses (e.g., Sections~\ref{sec:norm_analysis}, \ref{sec:layer_selection}), we focus our rigorous significance testing on this most conservative case. While we do not exhaustively bootstrap all architectures, the larger point estimates for Gemma 4 and LFM2-VL on the same large test set ($n=2,442$) strongly suggest their improvements are similarly robust. 

We assess significance via paired bootstrap resampling~\citep{efron1979bootstrap, efron1993introduction}, which estimates uncertainty directly from the data without assuming any particular score distribution. In each of 10{,}000 iterations, we draw test instances with replacement, compute the paired difference (ORBIT minus zero-shot) for each instance, and record the aggregated tier-level gain. The 2.5th and 97.5th percentiles of the resulting distribution form the 95\% confidence interval. As shown in Table~\ref{tab:bootstrap_low_resource}, all intervals for both accuracy and macro F1 exclude zero across every low-resource evidence language, confirming that ORBIT's recovery is statistically significant even under the weakest-model setting.

\section{Classification Report}
\label{appendix:classification_report}

\begin{table}[H]
\centering
\setlength{\tabcolsep}{6pt}
\begin{tabular}{lcccr}
\toprule
\textbf{Evidence Language} & \textbf{Supported} & \textbf{Refuted} & \textbf{NEI} & $\Delta$\textbf{Macro-F1} \\
\midrule
en & 1.47 & 2.01 & -8.38 & -1.63 \\
es & 1.31 & 2.68 & -7.25 & -1.09 \\
ru & 1.97 & 3.00 & -6.55 & -0.53 \\
vi & 1.93 & 3.61 & -6.88 & -0.44 \\
ja & 2.34 & 5.89 & -7.76 & 0.16 \\
\midrule
jw & 2.72 & 7.86 & -5.47 & 1.70 \\
cy & 2.07 & 7.35 & -3.89 & 1.84 \\
mi & 1.37 & 8.87 & -2.57 & 2.56 \\
ti & 1.40 & 7.05 & -0.43 & 2.68 \\
\bottomrule
\end{tabular}%
\caption{Class-wise $\Delta$F1 on Qwen3-VL by evidence language. Values report ORBIT minus zero-shot F1, averaged over all claim languages for each evidence language.}
\label{tab:evidence_delta_f1}
\end{table}

Table~\ref{tab:evidence_delta_f1} decomposes ORBIT's macro F1 changes on Qwen3-VL by individual class performance. F1 drops on high-resource evidence stem almost entirely from the NEI class, while Supported and Refuted verdicts improve modestly. Uniform steering overshoots well-aligned representations and collapses the ambiguity required for abstention. Conversely, low-resource evidence benefits across all three classes, as ORBIT pulls severely misaligned representations back toward viable decision boundaries.

Tables~\ref{tab:supported_ORBIT_heatmap}, \ref{tab:Refuted_ORBIT_heatmap}, and~\ref{tab:NEI_ORBIT_heatmap} show the full per-configuration breakdown for each class on Qwen3-VL. The pattern holds across all claim languages: high-resource evidence columns exhibit systematic NEI degradation, while low-resource columns show predominantly positive changes for Supported and Refuted, with NEI recovering from collapse.

\section{Calibration Size Sensitivity}
\label{app:calibration_sensitivity}

\begin{table}[H]
\centering
\resizebox{\linewidth}{!}{%
\setlength{\tabcolsep}{6pt}
\begin{tabular}{lcccccccc}
\toprule
\multirow{2}{*}{$N$} 
& \multicolumn{2}{c}{\textbf{vi--jw}} 
& \multicolumn{2}{c}{\textbf{vi--cy}} 
& \multicolumn{2}{c}{\textbf{vi--mi}} 
& \multicolumn{2}{c}{\textbf{vi--ti}} \\
\cmidrule(lr){2-3}\cmidrule(lr){4-5}\cmidrule(lr){6-7}\cmidrule(lr){8-9}
& Accuracy & F1-score & Accuracy & F1-score & Accuracy & F1-score & Accuracy & F1-score \\
\midrule
32  & 48.52 & 46.67 & 45.57 & 45.22 & 40.82 & 41.17 & 40.48 & 40.70 \\
64  & 48.34 & 46.51 & 45.30 & 44.96 & 40.82 & 41.17 & 40.66 & 40.85 \\
128 & 48.61 & 46.84 & 46.02 & 45.69 & 40.91 & 41.26 & 40.66 & 40.88 \\
256 & 48.52 & 46.76 & 45.75 & 45.44 & 41.00 & 41.34 & 40.48 & 40.65 \\
372 & 48.61 & 46.84 & 45.48 & 45.13 & 40.73 & 41.07 & 40.48 & 40.69 \\
\bottomrule
\end{tabular}
}
\caption{Calibration-size sensitivity for ORBIT on Qwen3-VL with Vietnamese claims and selected low-resource evidence languages.}
\label{tab:calibration_size}
\end{table}

We fix the claim language to Vietnamese and evaluate low-resource evidence settings on Qwen3-VL: vi--jw, vi--cy, vi--mi, and vi--ti. Vietnamese is chosen because it is a high-resource non-English language in our setup, providing a representative cross-lingual claim condition without relying on the English source language as the claim input. This also controls claim-side variation, so the analysis focuses on how calibration size affects steering under low-resource evidence shift. We vary the calibration size $N \in \{32,64,128,256,372\}$ while keeping all other hyperparameters fixed. Table~\ref{tab:calibration_size} reports the per-setting accuracy and macro F1 changes.





\section{Steering Direction}
\label{app:sign_ablation}


\begin{table}[h]
\centering
\small
\begin{tabular}{lcccccc}
\toprule
\multirow{2}{*}{\textbf{Method}} & \multicolumn{2}{c}{\textbf{cy--ru}} & \multicolumn{2}{c}{\textbf{mi--cy}} & \multicolumn{2}{c}{\textbf{vi--jw}} \\
\cmidrule(lr){2-3} \cmidrule(lr){4-5} \cmidrule(lr){6-7}
 & Accuracy & F1-score & Accuracy & F1-score & Accuracy & F1-score \\
\midrule
Zero-shot & 44.63 & 43.87 & 39.66 & 39.20 & 46.30 & 45.64 \\
Reversed ($h - \lambda v$) & 43.99 & 42.29 & 36.49 & 36.42 & 46.57 & 45.64 \\
\textbf{ORBIT ($h + \lambda v$)} & \textbf{46.81} & \textbf{45.13} & \textbf{39.85} & \textbf{39.74} & \textbf{49.48} & \textbf{47.45} \\
\bottomrule
\end{tabular}

\caption{Steering direction ablation on Qwen3-VL comparing additive realignment ($h + \lambda v$) against reversed steering ($h - \lambda v$) and the zero-shot baseline. Best result per pair and metric in bold.}
\label{tab:sign_ablation}
\end{table}

We show the impact of steering direction in Equations~\ref{eq:shift_claim}--\ref{eq:shift_evidence} and Equations~\ref{eq:steer_claim}--\ref{eq:steer_evidence} by ablating it on Qwen3-VL across three configurations (cy--ru, mi--cy, and vi--jw). In Table~\ref{tab:sign_ablation}, additive steering improves over the zero-shot baseline, whereas reversing the sign pushes target-language hidden states further away from the anchor space, resulting in a decline in accuracy. By contrast, ORBIT shows consistent improvements in both accuracy and F1-score across all three pairs.

\section{Control Experiments}
\label{app:control_experiments}

\begin{table}[h]
\centering
\small
\begin{tabular}{lcccccc}
\toprule
\multirow{2}{*}{\textbf{Method}} & \multicolumn{2}{c}{\textbf{Qwen3-VL}} & \multicolumn{2}{c}{\textbf{Gemma 4}} & \multicolumn{2}{c}{\textbf{LFM2-VL}} \\
\cmidrule(lr){2-3} \cmidrule(lr){4-5} \cmidrule(lr){6-7}
 & $\Delta$Acc & $\Delta$F1 & $\Delta$Acc & $\Delta$F1 & $\Delta$Acc & $\Delta$F1 \\
\midrule
Mismatched vector (es--ti) & -0.12 & -0.96 & -0.10 & -0.34 & +1.53 & +1.21 \\
Same-norm random vector & +0.50 & -0.19 & +0.70 & -0.14 & +2.41 & +1.51 \\
Label-bias calibration & -0.29 & +0.02 & +0.13 & +0.65 & +1.48 & +1.56 \\
\textbf{ORBIT} & \textbf{+1.98} & \textbf{+1.46} & \textbf{+3.19} & \textbf{+2.18} & \textbf{+5.76} & \textbf{+5.18} \\
\bottomrule
\end{tabular}

\caption{Mean $\Delta$ relative to zero-shot over the six claim--evidence pairs on the validation split. Full per-pair results are in Table~\ref{tab:control_ablation}. Best result per model and metric in bold.}
\label{tab:control_compact}
\end{table}

We validate that the improvement from ORBIT does not simply come from adding any vector or from changing how often each label is predicted. We test three controls on all three models and six pairs (cy--ru, mi--cy, vi--jw, mi--ti, ti--ja, ru--jw) -- covering low-resource claims, low-resource evidence, and both sides. The first control uses a mismatched vector which is calibrated from es--ti pair, while the second uses a random vector with the same norm as the ORBIT vector. Finally, the third is label-bias calibration, which learns three per-class offsets on the label logits from the 372 calibration samples so that the predicted label proportions match the balanced label prior, without using ground-truth labels.

Table~\ref{tab:control_ablation} shows mismatched vector is unstable, since it causes a drop in mi--cy and lowers F1-score on cy--ru for Qwen3-VL, but enhances LFM2-VL performance. The random vector slightly improves accuracy but hardly changes F1-score on Qwen3-VL and Gemma 4. Label-bias calibration also changes little on these two models and can even be worse than zero-shot. On LFM2-VL, all three controls improve, but averaged over the six pairs, ORBIT is still more than three points better than the best control on both metrics. Overall, ORBIT shows consistent improvements both in accuracy and F1-score on all six pairs and three models, showing that language-matched steering directions matter.

\begin{table}[h]
\centering
\small
\resizebox{\textwidth}{!}{%
\begin{tabular}{lcccccccccccc}
\toprule
\multirow{2}{*}{\textbf{Method}} & \multicolumn{2}{c}{\textbf{cy--ru}} & \multicolumn{2}{c}{\textbf{mi--cy}} & \multicolumn{2}{c}{\textbf{vi--jw}} & \multicolumn{2}{c}{\textbf{mi--ti}} & \multicolumn{2}{c}{\textbf{ti--ja}} & \multicolumn{2}{c}{\textbf{ru--jw}} \\
\cmidrule(lr){2-3} \cmidrule(lr){4-5} \cmidrule(lr){6-7} \cmidrule(lr){8-9} \cmidrule(lr){10-11} \cmidrule(lr){12-13}
 & Acc & F1 & Acc & F1 & Acc & F1 & Acc & F1 & Acc & F1 & Acc & F1 \\
\midrule
\multicolumn{13}{l}{\textbf{Qwen3-VL}} \\
Zero-shot & 44.63 & 43.87 & 39.66 & 39.20 & 46.30 & 45.64 & 35.19 & 33.55 & 37.61 & 36.44 & 46.51 & 45.85 \\
Mismatched vector (es--ti) & 44.63 & 42.99 & 36.49 & 36.41 & 48.84 & 46.95 & 34.44 & 29.66 & 36.94 & 37.00 & 47.82 & 45.77 \\
Same-norm random vector & 44.54 & 42.82 & 38.28 & 38.25 & 47.55 & 45.26 & 35.87 & 32.13 & 37.39 & 37.71 & 49.27 & 47.22 \\
Label-bias calibration & 44.99 & 43.19 & 39.25 & 39.31 & 45.91 & 45.79 & 35.21 & 33.18 & 36.62 & 36.59 & 46.15 & 46.63 \\
\textbf{ORBIT} & \textbf{46.81} & \textbf{45.13} & \textbf{39.85} & \textbf{39.74} & \textbf{49.48} & \textbf{47.45} & \textbf{37.35} & \textbf{35.50} & \textbf{38.94} & \textbf{38.06} & \textbf{49.37} & \textbf{47.43} \\
\midrule
\multicolumn{13}{l}{\textbf{Gemma 4}} \\
Zero-shot & 34.53 & 34.87 & 33.45 & 31.77 & 36.76 & 36.58 & 30.77 & 28.26 & 35.78 & 35.82 & 35.33 & 35.52 \\
Mismatched vector (es--ti) & 34.37 & 33.65 & 33.45 & 32.32 & 36.88 & 35.99 & 30.23 & 30.34 & 36.55 & 34.58 & 34.52 & 33.92 \\
Same-norm random vector & 37.22 & 35.22 & 32.46 & 32.72 & 37.02 & 35.44 & 30.65 & 29.13 & 37.43 & 34.59 & 36.06 & 34.91 \\
Label-bias calibration & 35.87 & 34.92 & 33.90 & 34.18 & 36.40 & 36.29 & 29.87 & 29.85 & 36.12 & 36.13 & 35.24 & 35.36 \\
\textbf{ORBIT} & \textbf{38.91} & \textbf{36.82} & \textbf{34.35} & \textbf{34.97} & \textbf{41.23} & \textbf{38.01} & \textbf{31.57} & \textbf{31.70} & \textbf{38.82} & \textbf{36.41} & \textbf{40.88} & \textbf{37.96} \\
\midrule
\multicolumn{13}{l}{\textbf{LFM2-VL}} \\
Zero-shot & 30.16 & 29.45 & 29.65 & 27.47 & 30.79 & 30.01 & 29.53 & 27.79 & 34.81 & 31.58 & 30.52 & 28.27 \\
Mismatched vector (es--ti) & 32.75 & 31.52 & 31.53 & 29.22 & 33.84 & 31.50 & 31.20 & 29.00 & 34.16 & 31.57 & 31.14 & 29.00 \\
Same-norm random vector & 35.01 & 32.86 & 32.88 & 28.50 & 31.92 & 30.86 & 30.20 & 28.21 & 35.70 & 34.00 & 34.23 & 29.22 \\
Label-bias calibration & 33.94 & 30.96 & 30.13 & 30.29 & 32.15 & 30.33 & 30.93 & 29.07 & 34.53 & 32.51 & 32.63 & 30.77 \\
\textbf{ORBIT} & \textbf{38.55} & \textbf{36.19} & \textbf{35.15} & \textbf{32.66} & \textbf{37.66} & \textbf{34.29} & \textbf{32.29} & \textbf{30.92} & \textbf{40.43} & \textbf{38.65} & \textbf{35.96} & \textbf{32.91} \\
\bottomrule
\end{tabular}%
}

\caption{Extended control experiments on the validation split across six claim--evidence pairs. Best result per pair and metric in bold.}
\label{tab:control_ablation}
\end{table}

\section{Component Analysis}
\label{app:component_ablation}

\begin{table}[h]
\centering
\small
\begin{tabular}{lcccccc}
\toprule
\multirow{2}{*}{\textbf{Method}} & \multicolumn{2}{c}{\textbf{cy--ru}} & \multicolumn{2}{c}{\textbf{mi--cy}} & \multicolumn{2}{c}{\textbf{vi--jw}} \\
\cmidrule(lr){2-3} \cmidrule(lr){4-5} \cmidrule(lr){6-7}
 & Accuracy & F1-score & Accuracy & F1-score & Accuracy & F1-score \\
\midrule
Zero-shot & 44.63 & 43.87 & 39.66 & 39.20 & 46.30 & 45.64 \\
\midrule
\multicolumn{7}{l}{\textit{Aggregation variants}} \\
Plain mean & 43.83 & 42.31 & 37.57 & 37.35 & 48.19 & 46.11 \\
Unit-normalized mean & 43.83 & 42.33 & 37.30 & 37.11 & 47.93 & 45.96 \\
\midrule
\multicolumn{7}{l}{\textit{Span weighting variants}} \\
Direct-norm weighting & 44.65 & 43.22 & 38.39 & 38.15 & 48.93 & 46.88 \\
Uniform weighting & 44.45 & 43.17 & 38.57 & 38.33 & 48.57 & 46.42 \\
\midrule
\textbf{ORBIT} & \textbf{46.81} & \textbf{45.13} & \textbf{39.85} & \textbf{39.74} & \textbf{49.48} & \textbf{47.45} \\
\bottomrule
\end{tabular}

\caption{Component ablations on Qwen3-VL evaluating aggregation strategies (Equation~\ref{eq:sign_vector}) and span weighting schemes (Equation~\ref{eq:adaptive_lambda}) against ORBIT and the zero-shot baseline.}
\label{tab:component_ablation}
\end{table}

We examine two design choices of ORBIT on Qwen3-VL across three representative configurations (cy--ru, mi--cy, and vi--jw): sign-aligned aggregation (Equation~\ref{eq:sign_vector}) and inverse-norm span weighting (Equation~\ref{eq:adaptive_lambda}). For aggregation, we compare against plain mean (directly averaging per-sample shifts) and unit-normalized mean (normalizing each sample shift to unit length before averaging). For span weighting, we compare against direct-norm weighting ($\lambda_c = \lambda \cdot \frac{n_c}{n_c + n_e}$, $\lambda_e = \lambda \cdot \frac{n_e}{n_c + n_e}$) and uniform weighting ($\lambda_c = \lambda_e = \lambda / 2$). Table~\ref{tab:component_ablation} shows that both components are necessary. Replacing sign alignment yields weaker results. Similarly, replacing inverse-norm weighting frequently drops both accuracy and F1-score below zero-shot, whereas ORBIT consistently improves across all pairs.

\section{Text-only ablation}
\label{app:text_only}


Table~\ref{tab:text_only_ablation} shows that performance degradation even without images input. On six representative cross-lingual pairs (cy--ru, mi--cy, vi--jw, mi--ti, ti--ja, ru--jw), accuracy falls by an average of 18.10, 11.67, and 14.39 points relative to text-only en--en on Qwen3-VL, Gemma 4, and LFM2-VL, respectively. Comparing these results with multimodal zero-shot baseline in Table~\ref{tab:control_ablation}, shows that incorporating visual evidence improves average accuracy across the same six pairs by +5.46 (Qwen3-VL), +0.66 (Gemma 4), and +2.81 (LFM2-VL) points.

However, translation noise alone cannot account for the collapse. On Qwen3-VL, cy--ru suffers a larger text-only accuracy drop (-15.79) than en--cy (-12.62), despite cy--ru having higher translation adequacy for both claim and evidence (shown in Table~\ref{tab:evaluation_scores}). Thus, low-resource collapse stems from both text comprehension failures and degraded cross-modal grounding, and this work focuses on the multimodal setting where both factors can play a role.

\begin{table}[h]
\centering
\small
\begin{tabular}{lcccccc}
\toprule
\multirow{2}{*}{\textbf{Pair}} & \multicolumn{2}{c}{\textbf{Qwen3-VL}} & \multicolumn{2}{c}{\textbf{Gemma 4}} & \multicolumn{2}{c}{\textbf{LFM2-VL}} \\
\cmidrule(lr){2-3} \cmidrule(lr){4-5} \cmidrule(lr){6-7}
 & Accuracy & F1-score & Accuracy & F1-score & Accuracy & F1-score \\
\midrule
\multicolumn{7}{l}{\textit{Text-only (no visual evidence)}} \\
en--en & 54.29 & 52.29 & 45.44 & 43.93 & 42.49 & 41.54 \\
en--cy & 41.67 & 42.45 & 34.83 & 33.32 & 26.16 & 19.95 \\
mi--en & 39.45 & 39.90 & 39.71 & 40.03 & 31.93 & 27.71 \\
cy--ru & 38.50 & 39.09 & 33.09 & 32.75 & 29.10 & 23.83 \\
mi--cy & 32.80 & 28.20 & 33.93 & 30.32 & 25.00 & 17.20 \\
vi--jw & 41.27 & 41.70 & 35.42 & 35.24 & 28.11 & 19.07 \\
mi--ti & 31.48 & 18.96 & 30.44 & 25.01 & 27.17 & 17.22 \\
ti--ja & 31.75 & 28.67 & 34.26 & 33.02 & 31.22 & 25.11 \\
ru--jw & 41.32 & 41.67 & 35.51 & 35.47 & 28.00 & 20.74 \\
\midrule
\multicolumn{7}{l}{\textit{$\Delta$ relative to text-only en--en}} \\
en--en & 0.00 & 0.00 & 0.00 & 0.00 & 0.00 & 0.00 \\
en--cy & -12.62 & -9.84 & -10.61 & -10.61 & -16.33 & -21.59 \\
mi--en & -14.84 & -12.39 & -5.73 & -3.90 & -10.56 & -13.83 \\
cy--ru & -15.79 & -13.20 & -12.35 & -11.18 & -13.39 & -17.71 \\
mi--cy & -21.49 & -24.09 & -11.51 & -13.61 & -17.49 & -24.34 \\
vi--jw & -13.02 & -10.59 & -10.02 & -8.69 & -14.38 & -22.47 \\
mi--ti & -22.81 & -33.33 & -15.00 & -18.92 & -15.32 & -24.32 \\
ti--ja & -22.54 & -23.62 & -11.18 & -10.91 & -11.27 & -16.43 \\
ru--jw & -12.97 & -10.62 & -9.93 & -8.46 & -14.49 & -20.80 \\
\bottomrule
\end{tabular}

\caption{Zero-shot text-only performance and degradation relative to en--en on the validation split across representative claim--evidence pairs.}
\label{tab:text_only_ablation}
\end{table}

\clearpage

\begin{figure}[H]
\centering
\tiny
\setlength{\tabcolsep}{2.75pt}
\setlength{\fboxsep}{1pt}

\begin{minipage}[t]{0.5\textwidth}
\centering
\renewcommand{\arraystretch}{1.15}
\begin{tabular}{l|ccccccccc}
\toprule
\multirow{2}{*}{\textbf{Claim}} & \multicolumn{9}{c}{\textbf{Evidence Language}} \\
\cmidrule{2-10}
& \textbf{en} & \textbf{es} & \textbf{ru} & \textbf{vi} & \textbf{ja} & \textbf{jw} & \textbf{cy} & \textbf{mi} & \textbf{ti} \\

\midrule
\textbf{en} & \nc{--} & \hc{+0.25} & \hc{+1.93} & \hc{+1.52} & \hc{+2.01} & \hc{+3.04} & \hc{+2.71} & \hc{+2.42} & \hc{+2.22} \\
\textbf{es} & \lc{-0.24} & \lc{0.00} & \hc{+1.28} & \hc{+1.11} & \hc{+1.39} & \hc{+3.07} & \hc{+2.18} & \hc{+1.97} & \hc{+2.79} \\
\textbf{ru} & \lc{-0.37} & \hc{+0.58} & \hc{+0.94} & \hc{+1.11} & \hc{+2.42} & \hc{+3.53} & \hc{+2.34} & \hc{+0.74} & \hc{+1.40} \\
\textbf{vi} & \lc{-0.65} & \hc{+1.28} & \hc{+1.32} & \hc{+1.97} & \hc{+3.00} & \hc{+3.56} & \hc{+2.34} & \hc{+2.83} & \hc{+2.01} \\
\textbf{ja} & \lc{-0.82} & \hc{+0.74} & \hc{+1.68} & \hc{+0.91} & \hc{+2.67} & \hc{+3.40} & \hc{+3.04} & \hc{+1.32} & \hc{+1.85} \\
\textbf{jw} & \hc{+1.35} & \hc{+2.05} & \hc{+2.95} & \hc{+2.33} & \hc{+3.69} & \hc{+2.71} & \hc{+1.68} & \hc{+2.50} & \hc{+1.93} \\
\textbf{cy} & \hc{+2.63} & \hc{+2.05} & \hc{+2.38} & \hc{+2.05} & \hc{+3.07} & \hc{+3.40} & \hc{+3.24} & \hc{+1.19} & \hc{+0.49} \\
\textbf{mi} & \hc{+1.19} & \hc{+0.50} & \hc{+0.49} & \hc{+0.62} & \hc{+0.74} & \hc{+0.58} & \hc{+0.90} & \hc{+1.52} & \hc{+0.12} \\
\textbf{ti} & \hc{+0.86} & \lc{-0.20} & \hc{+0.01} & \hc{+0.74} & \lc{-0.12} & \hc{+0.66} & \hc{+0.09} & \lc{0.00} & \hc{+2.05} \\
\bottomrule
\end{tabular}
\caption*{$\Delta$ Accuracy (\%)}
\end{minipage}%
\hfill
%
\begin{minipage}[t]{0.5\textwidth}
\centering
\renewcommand{\arraystretch}{1.15}
\begin{tabular}{l|ccccccccc}
\toprule
\multirow{2}{*}{\textbf{Claim}} & \multicolumn{9}{c}{\textbf{Evidence Language}} \\
\cmidrule{2-10}
& \textbf{en} & \textbf{es} & \textbf{ru} & \textbf{vi} & \textbf{ja} & \textbf{jw} & \textbf{cy} & \textbf{mi} & \textbf{ti} \\

\midrule
\textbf{en} & \nc{--} & \lc{-2.18} & \lc{-0.75} & \lc{-0.87} & \lc{-1.04} & \hc{+1.27} & \hc{+2.07} & \hc{+2.91} & \hc{+2.51} \\
\textbf{es} & \lc{-2.83} & \lc{-2.63} & \lc{-1.31} & \lc{-1.43} & \lc{-1.71} & \hc{+1.36} & \hc{+1.55} & \hc{+2.36} & \hc{+3.10} \\
\textbf{ru} & \lc{-3.27} & \lc{-2.39} & \lc{-2.12} & \lc{-1.58} & \lc{-0.68} & \hc{+1.46} & \hc{+1.75} & \hc{+1.07} & \hc{+1.84} \\
\textbf{vi} & \lc{-3.44} & \lc{-1.47} & \lc{-1.40} & \lc{-1.06} & \hc{+0.06} & \hc{+1.68} & \hc{+1.85} & \hc{+3.58} & \hc{+2.32} \\
\textbf{ja} & \lc{-3.95} & \lc{-2.30} & \lc{-1.42} & \lc{-2.05} & \lc{-1.00} & \hc{+1.32} & \hc{+2.40} & \hc{+1.92} & \hc{+2.44} \\
\textbf{jw} & \lc{-1.47} & \hc{+0.05} & \hc{+0.84} & \hc{+0.40} & \hc{+1.91} & \hc{+0.63} & \hc{+2.41} & \hc{+4.20} & \hc{+3.16} \\
\textbf{cy} & \hc{+0.64} & \hc{+0.92} & \hc{+1.29} & \hc{+1.19} & \hc{+2.51} & \hc{+4.43} & \hc{+1.82} & \hc{+2.48} & \hc{+1.20} \\
\textbf{mi} & \hc{+0.39} & \hc{+0.21} & \hc{+0.16} & \hc{+0.48} & \hc{+0.81} & \hc{+1.52} & \hc{+1.62} & \hc{+2.03} & \hc{+0.23} \\
\textbf{ti} & \hc{+0.87} & \hc{+0.00} & \lc{-0.05} & \hc{+0.93} & \hc{+0.54} & \hc{+1.11} & \hc{+0.07} & \hc{+0.05} & \hc{+3.60} \\
\bottomrule
\end{tabular}
\caption*{$\Delta$ F1-score (\%)}
\end{minipage}

\vspace{0.3cm}

\begin{minipage}{\textwidth}
\centering
\scriptsize
\textbf{Improvement levels:} 
\colorbox{highperf!30}{\strut~Positive ($>$0)~} 
\colorbox{lowperf!50}{\strut~Zero or Negative ($\leq$\,0)~} 
\end{minipage}

\caption{Per-configuration ORBIT gains ($\Delta$ = \textbf{ORBIT} $-$ \textbf{Zero-shot}) on Qwen3-VL across all 81 claim--evidence language pairs.}
\label{fig:orbit_delta_heatmap}
\end{figure}

\begin{figure}[H]
\centering
\tiny
\setlength{\tabcolsep}{2.75pt}
\setlength{\fboxsep}{1pt}

\begin{minipage}[t]{0.5\textwidth}
\centering
\renewcommand{\arraystretch}{1.15}
\begin{tabular}{l|ccccccccc}
\toprule
\multirow{2}{*}{\textbf{Claim}} & \multicolumn{9}{c}{\textbf{Evidence Language}} \\
\cmidrule{2-10}
& \textbf{en} & \textbf{es} & \textbf{ru} & \textbf{vi} & \textbf{ja} & \textbf{jw} & \textbf{cy} & \textbf{mi} & \textbf{ti} \\

\midrule
\textbf{en} & \nc{--} & \hc{+0.49} & \hc{+0.20} & \hc{+1.06} & \hc{+0.12} & \hc{+1.11} & \hc{+0.25} & \lc{-0.29} & \lc{-1.19} \\
\textbf{es} & \lc{-0.33} & \lc{-0.57} & \lc{-0.29} & \hc{+0.74} & \hc{+0.90} & \hc{+2.13} & \hc{+0.61} & \hc{+1.39} & \hc{+0.74} \\
\textbf{ru} & \lc{-0.82} & \lc{-0.25} & \lc{-0.49} & \hc{+0.70} & \hc{+2.25} & \hc{+3.77} & \hc{+0.90} & \hc{+0.16} & \hc{+0.78} \\
\textbf{vi} & \lc{-0.16} & \hc{+1.56} & \hc{+2.29} & \hc{+0.86} & \hc{+2.74} & \hc{+3.64} & \hc{+2.25} & \hc{+2.74} & \hc{+0.45} \\
\textbf{ja} & \lc{-1.64} & \hc{+1.56} & \hc{+0.70} & \lc{-0.53} & \hc{+1.64} & \hc{+3.32} & \hc{+1.11} & \lc{-0.41} & \hc{+0.16} \\
\textbf{jw} & \hc{+0.08} & \hc{+2.01} & \hc{+2.38} & \hc{+2.38} & \hc{+2.33} & \hc{+0.04} & \hc{+3.60} & \hc{+2.29} & \hc{+2.74} \\
\textbf{cy} & \lc{-2.66} & \lc{-1.19} & \lc{-0.37} & \lc{-1.02} & \hc{+0.37} & \hc{+2.70} & \lc{-3.52} & \hc{+0.37} & \hc{+0.33} \\
\textbf{mi} & \lc{-2.54} & \lc{-1.52} & \lc{-1.39} & \lc{-1.11} & \lc{-0.29} & \hc{+1.47} & \hc{+2.33} & \hc{+0.16} & \hc{+0.74} \\
\textbf{ti} & \lc{-6.06} & \lc{-5.16} & \lc{-2.62} & \lc{-3.77} & \lc{-2.54} & \lc{-1.02} & \lc{-1.80} & \lc{-0.82} & \lc{-2.99} \\
\bottomrule
\end{tabular}
\caption*{$\Delta$ Accuracy (\%)}
\end{minipage}%
\hfill
%
\begin{minipage}[t]{0.5\textwidth}
\centering
\renewcommand{\arraystretch}{1.15}
\begin{tabular}{l|ccccccccc}
\toprule
\multirow{2}{*}{\textbf{Claim}} & \multicolumn{9}{c}{\textbf{Evidence Language}} \\
\cmidrule{2-10}
& \textbf{en} & \textbf{es} & \textbf{ru} & \textbf{vi} & \textbf{ja} & \textbf{jw} & \textbf{cy} & \textbf{mi} & \textbf{ti} \\

\midrule
\textbf{en} & \nc{--} & \hc{+0.71} & \lc{-0.10} & \hc{+0.76} & \lc{-0.85} & \hc{+0.37} & \lc{-0.21} & \lc{-0.88} & \lc{-5.26} \\
\textbf{es} & \hc{+0.27} & \lc{-0.49} & \lc{-0.46} & \hc{+0.45} & \lc{-0.01} & \hc{+1.37} & \hc{+0.09} & \hc{+1.56} & \lc{-2.10} \\
\textbf{ru} & \lc{-1.91} & \lc{-2.28} & \lc{-2.27} & \lc{-1.27} & \lc{-0.39} & \hc{+1.67} & \lc{-0.01} & \hc{+0.04} & \lc{-2.24} \\
\textbf{vi} & \lc{-1.72} & \lc{-0.96} & \hc{+0.41} & \lc{-1.07} & \lc{-0.03} & \hc{+1.65} & \hc{+1.27} & \hc{+3.05} & \lc{-2.40} \\
\textbf{ja} & \lc{-2.48} & \hc{+0.08} & \lc{-1.13} & \lc{-2.14} & \lc{-0.70} & \hc{+1.73} & \hc{+0.38} & \lc{-0.41} & \lc{-2.76} \\
\textbf{jw} & \lc{-0.89} & \hc{+0.51} & \hc{+0.70} & \hc{+0.80} & \hc{+0.28} & \lc{-1.20} & \hc{+3.54} & \hc{+3.17} & \hc{+0.28} \\
\textbf{cy} & \lc{-3.09} & \lc{-2.17} & \lc{-1.43} & \lc{-2.02} & \lc{-1.15} & \hc{+2.69} & \lc{-4.06} & \hc{+0.94} & \lc{-2.04} \\
\textbf{mi} & \lc{-3.68} & \lc{-2.70} & \lc{-2.70} & \lc{-2.16} & \lc{-1.51} & \hc{+0.90} & \hc{+1.83} & \lc{-1.52} & \lc{-1.79} \\
\textbf{ti} & \lc{-12.12} & \lc{-10.90} & \lc{-8.04} & \lc{-9.32} & \lc{-6.91} & \lc{-5.60} & \lc{-6.17} & \lc{-4.38} & \lc{-8.34} \\
\bottomrule
\end{tabular}
\caption*{$\Delta$ F1-score (\%)}
\end{minipage}

\vspace{0.3cm}

\begin{minipage}{\textwidth}
\centering
\scriptsize
\textbf{Improvement levels:} 
\colorbox{highperf!30}{\strut~Positive ($>$0)~} 
\colorbox{lowperf!50}{\strut~Zero or Negative ($\leq$\,0)~} 
\end{minipage}

\caption{Per-configuration Self-translate CoT gains ($\Delta$ = \textbf{Self-translate CoT} $-$ \textbf{Zero-shot}) on Qwen3-VL across all 81 claim--evidence language pairs.}
\label{fig:cot_delta_heatmap}
\end{figure}

\begin{figure}[H]
\centering
\tiny
\setlength{\tabcolsep}{2.75pt}
\setlength{\fboxsep}{1pt}

\begin{minipage}[t]{0.5\textwidth}
\centering
\renewcommand{\arraystretch}{1.15}
\begin{tabular}{l|ccccccccc}
\toprule
\multirow{2}{*}{\textbf{Claim}} & \multicolumn{9}{c}{\textbf{Evidence Language}} \\
\cmidrule{2-10}
& \textbf{en} & \textbf{es} & \textbf{ru} & \textbf{vi} & \textbf{ja} & \textbf{jw} & \textbf{cy} & \textbf{mi} & \textbf{ti} \\

\midrule
\textbf{en} & \nc{--} & \lc{-0.29} & \lc{-0.04} & \hc{+0.61} & \hc{+0.98} & \hc{+2.21} & \lc{-0.78} & \hc{+0.45} & \hc{+4.56} \\
\textbf{es} & \hc{+0.53} & \lc{-1.88} & \lc{-0.78} & \lc{-1.06} & \hc{+1.06} & \hc{+2.33} & \lc{-0.90} & \hc{+1.43} & \hc{+5.13} \\
\textbf{ru} & \lc{-0.74} & \lc{-2.62} & \lc{-2.33} & \lc{-1.97} & \hc{+0.82} & \hc{+1.97} & \lc{-1.27} & \hc{+0.16} & \hc{+4.44} \\
\textbf{vi} & \lc{-1.56} & \lc{-2.46} & \lc{-1.23} & \lc{-1.39} & \hc{+0.45} & \hc{+0.70} & \lc{-0.70} & \hc{+3.19} & \hc{+3.94} \\
\textbf{ja} & \lc{-1.68} & \lc{-1.56} & \hc{+0.20} & \lc{-1.84} & \hc{+0.25} & \hc{+1.39} & \hc{+0.04} & \lc{-0.37} & \hc{+3.62} \\
\textbf{jw} & \hc{+0.12} & \hc{+0.66} & \hc{+2.01} & \hc{+0.33} & \hc{+2.38} & \lc{-0.82} & \hc{+1.39} & \hc{+2.46} & \hc{+3.62} \\
\textbf{cy} & \lc{-4.79} & \lc{-4.42} & \lc{-2.09} & \lc{-3.81} & \lc{-1.06} & \hc{+1.64} & \lc{-5.53} & \hc{+0.29} & \lc{-0.81} \\
\textbf{mi} & \lc{-5.16} & \lc{-5.00} & \lc{-3.24} & \lc{-5.24} & \lc{-2.74} & \lc{-1.47} & \lc{-2.13} & \lc{-1.97} & \lc{-1.91} \\
\textbf{ti} & \hc{+1.45} & \lc{-0.36} & \hc{+2.96} & \hc{+0.67} & \hc{+3.37} & \hc{+1.00} & \lc{-1.67} & \lc{-1.83} & \hc{+1.16} \\
\bottomrule
\end{tabular}
\caption*{$\Delta$ Accuracy (\%)}
\end{minipage}%
\hfill
%
\begin{minipage}[t]{0.5\textwidth}
\centering
\renewcommand{\arraystretch}{1.15}
\begin{tabular}{l|ccccccccc}
\toprule
\multirow{2}{*}{\textbf{Claim}} & \multicolumn{9}{c}{\textbf{Evidence Language}} \\
\cmidrule{2-10}
& \textbf{en} & \textbf{es} & \textbf{ru} & \textbf{vi} & \textbf{ja} & \textbf{jw} & \textbf{cy} & \textbf{mi} & \textbf{ti} \\

\midrule
\textbf{en} & \nc{--} & \lc{-2.35} & \lc{-2.63} & \lc{-1.51} & \lc{-0.63} & \hc{+0.31} & \lc{-1.64} & \lc{-1.40} & \hc{+4.37} \\
\textbf{es} & \lc{-1.28} & \lc{-3.97} & \lc{-2.95} & \lc{-2.95} & \lc{-0.46} & \hc{+0.45} & \lc{-2.00} & \lc{-0.79} & \hc{+4.86} \\
\textbf{ru} & \lc{-2.38} & \lc{-4.78} & \lc{-4.39} & \lc{-3.65} & \lc{-0.97} & \lc{-0.09} & \lc{-2.64} & \lc{-2.18} & \hc{+3.80} \\
\textbf{vi} & \lc{-3.24} & \lc{-4.61} & \lc{-3.03} & \lc{-2.66} & \lc{-1.22} & \lc{-1.09} & \lc{-1.77} & \hc{+0.88} & \hc{+3.61} \\
\textbf{ja} & \lc{-2.85} & \lc{-3.34} & \lc{-1.79} & \lc{-3.36} & \lc{-1.20} & \lc{-0.27} & \lc{-1.23} & \lc{-2.95} & \hc{+3.17} \\
\textbf{jw} & \lc{-1.72} & \lc{-1.41} & \lc{-0.04} & \lc{-1.56} & \lc{-0.05} & \lc{-2.31} & \hc{+0.62} & \hc{+0.67} & \hc{+3.64} \\
\textbf{cy} & \lc{-6.37} & \lc{-6.61} & \lc{-4.31} & \lc{-5.87} & \lc{-4.28} & \hc{+0.53} & \lc{-7.50} & \lc{-2.77} & \hc{+1.65} \\
\textbf{mi} & \lc{-6.96} & \lc{-7.40} & \lc{-5.60} & \lc{-7.75} & \lc{-7.08} & \lc{-3.06} & \lc{-4.67} & \lc{-6.67} & \hc{+1.98} \\
\textbf{ti} & \hc{+0.14} & \lc{-1.33} & \hc{+2.16} & \lc{-0.18} & \hc{+2.41} & \hc{+1.91} & \hc{+1.64} & \lc{-0.96} & \hc{+0.36} \\
\bottomrule
\end{tabular}
\caption*{$\Delta$ F1-score (\%)}
\end{minipage}

\vspace{0.3cm}

\begin{minipage}{\textwidth}
\centering
\scriptsize
\textbf{Improvement levels:} 
\colorbox{highperf!30}{\strut~Positive ($>$0)~} 
\colorbox{lowperf!50}{\strut~Zero or Negative ($\leq$\,0)~} 
\end{minipage}

\caption{Per-configuration Pivot NLLB gains ($\Delta$ = \textbf{Pivot NLLB} $-$ \textbf{Zero-shot}) on Qwen3-VL across all 81 claim--evidence language pairs.}
\label{fig:nllb_delta_heatmap}
\end{figure}

\begin{figure}[H]
\centering
\tiny
\setlength{\tabcolsep}{2.75pt}
\setlength{\fboxsep}{1pt}

\begin{minipage}[t]{0.5\textwidth}
\centering
\renewcommand{\arraystretch}{1.15}
\begin{tabular}{l|ccccccccc}
\toprule
\multirow{2}{*}{\textbf{Claim}} & \multicolumn{9}{c}{\textbf{Evidence Language}} \\
\cmidrule{2-10}
& \textbf{en} & \textbf{es} & \textbf{ru} & \textbf{vi} & \textbf{ja} & \textbf{jw} & \textbf{cy} & \textbf{mi} & \textbf{ti} \\

\midrule
\textbf{en} & \nc{--} & \lc{-0.53} & \lc{-0.13} & \hc{+0.73} & \hc{+0.39} & \lc{-6.31} & \lc{-5.41} & \hc{+0.98} & \lc{-0.95} \\
\textbf{es} & \lc{-0.30} & \lc{-1.27} & \lc{-1.36} & \lc{-0.09} & \hc{+1.10} & \hc{+1.88} & \hc{+1.38} & \hc{+1.96} & \hc{+1.97} \\
\textbf{ru} & \lc{-1.39} & \lc{-0.98} & \lc{-0.58} & \hc{+0.57} & \hc{+1.63} & \hc{+2.86} & \hc{+2.29} & \hc{+0.97} & \hc{+1.84} \\
\textbf{vi} & \lc{-1.76} & \lc{0.00} & \lc{-0.58} & \lc{-0.09} & \hc{+0.85} & \hc{+2.29} & \hc{+0.54} & \hc{+3.39} & \hc{+1.23} \\
\textbf{ja} & \lc{-0.90} & \lc{-0.12} & \hc{+0.12} & \hc{+0.24} & \hc{+1.88} & \hc{+1.47} & \hc{+0.97} & \hc{+1.06} & \hc{+1.47} \\
\textbf{jw} & \lc{-0.04} & \hc{+0.24} & \hc{+1.56} & \hc{+0.98} & \hc{+1.55} & \hc{+2.08} & \hc{+1.14} & \hc{+3.40} & \hc{+2.20} \\
\textbf{cy} & \hc{+1.02} & \hc{+0.94} & \hc{+1.59} & \hc{+1.23} & \hc{+1.22} & \hc{+1.92} & \hc{+2.20} & \hc{+1.80} & \hc{+0.57} \\
\textbf{mi} & \hc{+1.43} & \hc{+0.74} & \hc{+0.33} & \hc{+0.77} & \hc{+0.45} & \hc{+0.65} & \hc{+0.16} & \hc{+2.13} & \hc{+0.28} \\
\textbf{ti} & \hc{+2.00} & \hc{+0.69} & \lc{-0.46} & \hc{+0.61} & \lc{-0.78} & \hc{+0.03} & \lc{-0.87} & \hc{+0.90} & \lc{-0.58} \\
\bottomrule
\end{tabular}
\caption*{$\Delta$ Accuracy (\%)}
\end{minipage}%
\hfill
%
\begin{minipage}[t]{0.5\textwidth}
\centering
\renewcommand{\arraystretch}{1.15}
\begin{tabular}{l|ccccccccc}
\toprule
\multirow{2}{*}{\textbf{Claim}} & \multicolumn{9}{c}{\textbf{Evidence Language}} \\
\cmidrule{2-10}
& \textbf{en} & \textbf{es} & \textbf{ru} & \textbf{vi} & \textbf{ja} & \textbf{jw} & \textbf{cy} & \textbf{mi} & \textbf{ti} \\

\midrule
\textbf{en} & \nc{--} & \lc{-1.14} & \lc{-1.17} & \lc{-0.28} & \lc{-1.32} & \lc{-7.46} & \lc{-6.25} & \hc{+0.85} & \lc{-1.64} \\
\textbf{es} & \lc{-1.98} & \lc{-2.56} & \lc{-2.88} & \lc{-1.62} & \lc{-0.92} & \hc{+0.48} & \hc{+0.37} & \hc{+1.97} & \hc{+2.92} \\
\textbf{ru} & \lc{-3.32} & \lc{-2.68} & \lc{-2.11} & \lc{-1.13} & \lc{-0.38} & \hc{+1.36} & \hc{+1.55} & \hc{+1.17} & \hc{+2.80} \\
\textbf{vi} & \lc{-3.63} & \lc{-1.50} & \lc{-2.15} & \lc{-1.57} & \lc{-1.02} & \hc{+0.85} & \lc{-0.02} & \hc{+3.91} & \hc{+1.67} \\
\textbf{ja} & \lc{-2.57} & \lc{-1.70} & \lc{-1.24} & \lc{-1.24} & \lc{-0.17} & \hc{+0.36} & \hc{+0.44} & \hc{+1.53} & \hc{+2.56} \\
\textbf{jw} & \lc{-2.18} & \lc{-1.66} & \lc{-0.47} & \lc{-0.80} & \lc{-0.12} & \hc{+0.33} & \hc{+1.25} & \hc{+5.00} & \hc{+3.89} \\
\textbf{cy} & \lc{-1.03} & \lc{-0.91} & \lc{-0.52} & \lc{-0.32} & \lc{-0.02} & \hc{+2.31} & \hc{+1.09} & \hc{+3.76} & \hc{+2.86} \\
\textbf{mi} & \lc{-0.41} & \lc{-0.71} & \lc{-1.24} & \lc{-0.34} & \lc{-0.21} & \hc{+1.26} & \hc{+0.98} & \hc{+2.32} & \hc{+2.08} \\
\textbf{ti} & \hc{+1.46} & \hc{+0.36} & \lc{-0.84} & \hc{+0.50} & \lc{-0.01} & \hc{+0.69} & \lc{-0.45} & \hc{+3.01} & \lc{-1.18} \\
\bottomrule
\end{tabular}
\caption*{$\Delta$ F1-score (\%)}
\end{minipage}

\vspace{0.3cm}

\begin{minipage}{\textwidth}
\centering
\scriptsize
\textbf{Improvement levels:} 
\colorbox{highperf!30}{\strut~Positive ($>$0)~} 
\colorbox{lowperf!50}{\strut~Zero or Negative ($\leq$\,0)~} 
\end{minipage}

\caption{Per-configuration CLAIM gains ($\Delta$ = \textbf{CLAIM} $-$ \textbf{Zero-shot}) on Qwen3-VL across all 81 claim--evidence language pairs.}
\label{fig:claim_delta_heatmap}
\end{figure}
\begin{figure}[H]
\centering
\tiny
\setlength{\tabcolsep}{2.75pt}
\setlength{\fboxsep}{1pt}

\begin{minipage}[t]{0.5\textwidth}
\centering
\renewcommand{\arraystretch}{1.15}
\begin{tabular}{l|ccccccccc}
\toprule
\multirow{2}{*}{\textbf{Claim}} & \multicolumn{9}{c}{\textbf{Evidence Language}} \\
\cmidrule{2-10}
& \textbf{en} & \textbf{es} & \textbf{ru} & \textbf{vi} & \textbf{ja} & \textbf{jw} & \textbf{cy} & \textbf{mi} & \textbf{ti} \\

\midrule
\textbf{en} & \nc{--} & \hc{+1.92} & \hc{+2.62} & \hc{+2.50} & \hc{+2.25} & \hc{+3.40} & \hc{+1.39} & \hc{+1.56} & \hc{+0.90} \\
\textbf{es} & \hc{+0.41} & \hc{+1.15} & \hc{+1.35} & \hc{+1.68} & \hc{+2.33} & \hc{+2.42} & \hc{+1.97} & \hc{+2.25} & \hc{+1.47} \\
\textbf{ru} & \hc{+1.19} & \hc{+1.76} & \hc{+1.92} & \hc{+1.39} & \hc{+2.01} & \hc{+2.25} & \hc{+1.68} & \hc{+1.15} & \hc{+1.68} \\
\textbf{vi} & \hc{+1.84} & \hc{+1.72} & \hc{+2.38} & \hc{+1.43} & \hc{+2.21} & \hc{+2.46} & \hc{+2.29} & \hc{+2.70} & \hc{+0.57} \\
\textbf{ja} & \hc{+1.39} & \hc{+2.01} & \hc{+3.48} & \hc{+2.95} & \hc{+0.94} & \hc{+2.87} & \hc{+1.72} & \hc{+1.76} & \hc{+2.25} \\
\textbf{jw} & \hc{+2.33} & \hc{+3.03} & \hc{+2.66} & \hc{+3.81} & \hc{+2.54} & \hc{+3.28} & \hc{+0.20} & \lc{-0.41} & \hc{+0.12} \\
\textbf{cy} & \hc{+2.29} & \hc{+5.00} & \hc{+3.15} & \hc{+4.14} & \hc{+2.58} & \hc{+2.05} & \hc{+3.15} & \hc{+0.86} & \hc{+0.41} \\
\textbf{mi} & \hc{+2.38} & \hc{+3.40} & \hc{+2.42} & \hc{+3.60} & \hc{+3.24} & \hc{+2.62} & \hc{+0.16} & \hc{+2.58} & \hc{+0.74} \\
\textbf{ti} & \hc{+0.04} & \hc{+3.32} & \hc{+2.87} & \hc{+2.01} & \hc{+2.78} & \hc{+0.70} & \hc{+0.49} & \hc{+0.08} & \hc{+3.89} \\

\bottomrule
\end{tabular}
\caption*{$\Delta$ Accuracy (\%)}
\end{minipage}%
\hfill
%
\begin{minipage}[t]{0.5\textwidth}
\centering
\renewcommand{\arraystretch}{1.15}
\begin{tabular}{l|ccccccccc}
\toprule
\multirow{2}{*}{\textbf{Claim}} & \multicolumn{9}{c}{\textbf{Evidence Language}} \\
\cmidrule{2-10}
& \textbf{en} & \textbf{es} & \textbf{ru} & \textbf{vi} & \textbf{ja} & \textbf{jw} & \textbf{cy} & \textbf{mi} & \textbf{ti} \\

\midrule
\textbf{en} & \nc{--} & \hc{+0.06} & \hc{+0.52} & \hc{+0.09} & \hc{+0.12} & \hc{+3.56} & \hc{+2.28} & \hc{+1.96} & \hc{+3.35} \\
\textbf{es} & \lc{-1.33} & \lc{-1.01} & \lc{-1.33} & \lc{-1.18} & \lc{-0.07} & \hc{+2.33} & \hc{+2.58} & \hc{+2.87} & \hc{+4.48} \\
\textbf{ru} & \lc{-0.01} & \lc{-0.50} & \lc{-1.29} & \lc{-1.19} & \lc{-0.08} & \hc{+2.31} & \hc{+2.89} & \hc{+2.40} & \hc{+4.69} \\
\textbf{vi} & \hc{+0.35} & \lc{-0.13} & \lc{-0.31} & \lc{-1.76} & \lc{-0.09} & \hc{+2.79} & \hc{+4.07} & \hc{+3.78} & \hc{+4.25} \\
\textbf{ja} & \hc{+0.58} & \hc{+0.30} & \hc{+1.04} & \hc{+0.88} & \lc{-2.11} & \hc{+3.57} & \hc{+3.48} & \hc{+3.26} & \hc{+5.71} \\
\textbf{jw} & \hc{+1.27} & \hc{+1.84} & \hc{+0.94} & \hc{+2.03} & \hc{+0.97} & \hc{+1.38} & \hc{+1.90} & \hc{+0.77} & \hc{+3.26} \\
\textbf{cy} & \hc{+2.09} & \hc{+4.92} & \hc{+2.44} & \hc{+3.37} & \hc{+2.05} & \hc{+3.32} & \lc{-0.22} & \hc{+2.50} & \hc{+3.38} \\
\textbf{mi} & \hc{+1.94} & \hc{+2.81} & \hc{+0.78} & \hc{+1.95} & \hc{+2.00} & \hc{+3.59} & \hc{+1.56} & \lc{-0.08} & \hc{+3.58} \\
\textbf{ti} & \lc{-0.24} & \hc{+3.42} & \hc{+2.36} & \hc{+1.65} & \hc{+2.43} & \hc{+2.94} & \hc{+3.05} & \hc{+2.88} & \hc{+3.78} \\
\bottomrule
\end{tabular}
\caption*{$\Delta$ F1-score (\%)}
\end{minipage}

\vspace{0.3cm}

\begin{minipage}{\textwidth}
\centering
\scriptsize
\textbf{Improvement levels:} 
\colorbox{highperf!30}{\strut~Positive ($>$0)~} 
\colorbox{lowperf!50}{\strut~Zero or Negative ($\leq$\,0)~} 
\end{minipage}

\caption{Per-configuration ORBIT gains ($\Delta$ = \textbf{ORBIT} $-$ \textbf{Zero-shot}) on Gemma 4 across all 81 claim--evidence language pairs.}
\label{fig:orbit_gemma4_delta_heatmap}
\end{figure}

\begin{figure}[H]
\centering
\tiny
\setlength{\tabcolsep}{2.75pt}
\setlength{\fboxsep}{1pt}

\begin{minipage}[t]{0.5\textwidth}
\centering
\renewcommand{\arraystretch}{1.15}
\begin{tabular}{l|ccccccccc}
\toprule
\multirow{2}{*}{\textbf{Claim}} & \multicolumn{9}{c}{\textbf{Evidence Language}} \\
\cmidrule{2-10}
& \textbf{en} & \textbf{es} & \textbf{ru} & \textbf{vi} & \textbf{ja} & \textbf{jw} & \textbf{cy} & \textbf{mi} & \textbf{ti} \\

\midrule
\textbf{en} & \nc{--} & \hc{+1.29} & \hc{+1.95} & \hc{+1.78} & \hc{+1.70} & \hc{+2.18} & \lc{-0.65} & \hc{+0.42} & \hc{+1.15} \\
\textbf{es} & \lc{-3.05} & \lc{-1.08} & \hc{+1.09} & \lc{-0.02} & \hc{+0.80} & \hc{+1.32} & \hc{+0.09} & \hc{+0.33} & \hc{+1.07} \\
\textbf{ru} & \lc{-2.23} & \hc{+0.39} & \hc{+1.01} & \hc{+0.97} & \hc{+1.25} & \hc{+1.36} & \hc{+0.29} & \hc{+0.46} & \hc{+1.36} \\
\textbf{vi} & \lc{-0.71} & \hc{+1.58} & \hc{+0.97} & \lc{-0.43} & \hc{+0.60} & \hc{+1.40} & \hc{+0.70} & \hc{+1.89} & \hc{+2.18} \\
\textbf{ja} & \lc{-0.67} & \lc{-0.18} & \hc{+1.78} & \hc{+1.21} & \hc{+0.11} & \hc{+2.10} & \hc{+1.44} & \hc{+2.34} & \hc{+2.30} \\
\textbf{jw} & \hc{+0.47} & \hc{+2.89} & \hc{+2.44} & \hc{+3.38} & \hc{+3.46} & \lc{-1.51} & \hc{+2.55} & \hc{+1.52} & \hc{+2.83} \\
\textbf{cy} & \lc{-2.39} & \hc{+0.84} & \hc{+0.11} & \hc{+1.42} & \hc{+0.19} & \lc{-1.80} & \lc{-3.56} & \lc{-1.59} & \lc{-0.73} \\
\textbf{mi} & \lc{-1.86} & \hc{+1.87} & \hc{+0.43} & \lc{-0.26} & \lc{-0.30} & \hc{+0.54} & \hc{+0.42} & \lc{-3.84} & \lc{-0.08} \\
\textbf{ti} & \lc{-4.52} & \lc{-0.63} & \lc{-1.29} & \lc{-1.57} & \lc{-1.16} & \lc{-1.88} & \lc{-0.20} & \lc{-0.98} & \lc{-3.47} \\

\bottomrule
\end{tabular}
\caption*{$\Delta$ Accuracy (\%)}
\end{minipage}%
\hfill
%
\begin{minipage}[t]{0.5\textwidth}
\centering
\renewcommand{\arraystretch}{1.15}
\begin{tabular}{l|ccccccccc}
\toprule
\multirow{2}{*}{\textbf{Claim}} & \multicolumn{9}{c}{\textbf{Evidence Language}} \\
\cmidrule{2-10}
& \textbf{en} & \textbf{es} & \textbf{ru} & \textbf{vi} & \textbf{ja} & \textbf{jw} & \textbf{cy} & \textbf{mi} & \textbf{ti} \\

\midrule
\textbf{en} & \nc{--} & \hc{+0.85} & \hc{+0.98} & \hc{+1.33} & \hc{+1.00} & \hc{+2.69} & \hc{+0.74} & \hc{+1.76} & \hc{+2.80} \\
\textbf{es} & \lc{-3.07} & \lc{-0.84} & \lc{-0.08} & \lc{-1.38} & \lc{-0.23} & \hc{+1.31} & \hc{+1.04} & \hc{+1.46} & \hc{+2.81} \\
\textbf{ru} & \lc{-1.99} & \lc{-0.08} & \hc{+0.98} & \hc{+0.06} & \hc{+0.49} & \hc{+1.81} & \hc{+1.89} & \hc{+1.87} & \hc{+3.02} \\
\textbf{vi} & \lc{-0.63} & \hc{+0.64} & \lc{-0.35} & \lc{-0.60} & \lc{-0.15} & \hc{+1.94} & \hc{+2.53} & \hc{+3.37} & \hc{+5.03} \\
\textbf{ja} & \lc{-0.09} & \lc{-0.48} & \hc{+0.96} & \hc{+0.64} & \lc{-0.14} & \hc{+2.74} & \hc{+3.23} & \hc{+3.83} & \hc{+4.47} \\
\textbf{jw} & \hc{+0.09} & \hc{+2.10} & \hc{+1.70} & \hc{+2.61} & \hc{+2.95} & \lc{-0.96} & \hc{+4.29} & \hc{+2.99} & \hc{+4.98} \\
\textbf{cy} & \lc{-2.28} & \hc{+0.84} & \hc{+0.22} & \hc{+1.48} & \hc{+0.36} & \lc{-1.14} & \lc{-3.51} & \lc{-0.71} & \lc{-0.20} \\
\textbf{mi} & \lc{-1.66} & \hc{+1.89} & \hc{+0.49} & \lc{-0.19} & \lc{-0.11} & \hc{+1.61} & \hc{+1.35} & \lc{-3.62} & \hc{+0.86} \\
\textbf{ti} & \lc{-4.85} & \lc{-1.02} & \lc{-1.39} & \lc{-1.64} & \lc{-1.39} & \lc{-1.01} & \hc{+0.70} & \hc{+0.50} & \lc{-4.55} \\

\bottomrule
\end{tabular}
\caption*{$\Delta$ F1-score (\%)}
\end{minipage}

\vspace{0.3cm}

\begin{minipage}{\textwidth}
\centering
\scriptsize
\textbf{Improvement levels:} 
\colorbox{highperf!30}{\strut~Positive ($>$0)~} 
\colorbox{lowperf!50}{\strut~Zero or Negative ($\leq$\,0)~} 
\end{minipage}

\caption{Per-configuration Self-translate CoT gains ($\Delta$ = \textbf{Self-translate CoT} $-$ \textbf{Zero-shot}) on Gemma 4 across all 81 claim--evidence language pairs.}
\label{fig:cot_gemma4_delta_heatmap}
\end{figure}

\begin{figure}[H]
\centering
\tiny
\setlength{\tabcolsep}{2.75pt}
\setlength{\fboxsep}{1pt}

\begin{minipage}[t]{0.5\textwidth}
\centering
\renewcommand{\arraystretch}{1.15}
\begin{tabular}{l|ccccccccc}
\toprule
\multirow{2}{*}{\textbf{Claim}} & \multicolumn{9}{c}{\textbf{Evidence Language}} \\
\cmidrule{2-10}
& \textbf{en} & \textbf{es} & \textbf{ru} & \textbf{vi} & \textbf{ja} & \textbf{jw} & \textbf{cy} & \textbf{mi} & \textbf{ti} \\

\midrule
\textbf{en} & \nc{--} & \lc{-0.29} & \hc{+1.88} & \hc{+0.45} & \hc{+1.64} & \hc{+2.24} & \lc{-0.08} & \lc{-1.11} & \hc{+4.15} \\
\textbf{es} & \lc{-2.01} & \lc{-2.54} & \lc{0.00} & \lc{-0.29} & \hc{+1.35} & \hc{+1.74} & \hc{+0.49} & \hc{+0.20} & \hc{+4.56} \\
\textbf{ru} & \lc{-1.35} & \lc{-0.25} & \hc{+0.12} & \hc{+0.41} & \hc{+1.68} & \hc{+1.33} & \hc{+1.64} & \hc{+1.39} & \hc{+4.39} \\
\textbf{vi} & \lc{-1.47} & \hc{+0.20} & \hc{+1.02} & \lc{-2.38} & \hc{+0.20} & \hc{+1.13} & \hc{+2.33} & \hc{+0.20} & \hc{+4.56} \\
\textbf{ja} & \lc{-1.64} & \lc{-0.66} & \hc{+2.01} & \hc{+1.72} & \hc{+0.78} & \hc{+2.73} & \hc{+2.95} & \hc{+1.88} & \hc{+5.01} \\
\textbf{jw} & \hc{+0.33} & \hc{+3.32} & \hc{+1.92} & \hc{+2.78} & \hc{+3.97} & \lc{-0.63} & \hc{+2.95} & \hc{+0.94} & \hc{+5.01} \\
\textbf{cy} & \lc{-1.60} & \hc{+0.49} & \hc{+0.57} & \hc{+1.15} & \hc{+1.60} & \hc{+0.11} & \lc{-3.24} & \lc{-0.45} & \hc{+0.95} \\
\textbf{mi} & \lc{-3.48} & \lc{-0.90} & \lc{-0.12} & \lc{-0.45} & \lc{-1.02} & \hc{+0.02} & \hc{+0.29} & \lc{-4.55} & \hc{+0.42} \\
\textbf{ti} & \hc{+0.66} & \hc{+1.52} & \hc{+2.01} & \hc{+0.25} & \hc{+2.79} & \hc{+1.11} & \lc{-0.16} & \lc{-0.94} & \lc{-0.03} \\

\bottomrule
\end{tabular}
\caption*{$\Delta$ Accuracy (\%)}
\end{minipage}%
\hfill
%
\begin{minipage}[t]{0.5\textwidth}
\centering
\renewcommand{\arraystretch}{1.15}
\begin{tabular}{l|ccccccccc}
\toprule
\multirow{2}{*}{\textbf{Claim}} & \multicolumn{9}{c}{\textbf{Evidence Language}} \\
\cmidrule{2-10}
& \textbf{en} & \textbf{es} & \textbf{ru} & \textbf{vi} & \textbf{ja} & \textbf{jw} & \textbf{cy} & \textbf{mi} & \textbf{ti} \\

\midrule
\textbf{en} & \nc{--} & \hc{+0.19} & \hc{+2.15} & \hc{+0.89} & \hc{+2.75} & \hc{+3.25} & \lc{-0.05} & \lc{-2.62} & \hc{+6.05} \\
\textbf{es} & \lc{-1.89} & \lc{-1.46} & \lc{-0.42} & \lc{-0.72} & \hc{+1.79} & \hc{+2.20} & \hc{+0.05} & \lc{-1.58} & \hc{+6.84} \\
\textbf{ru} & \lc{-1.01} & \lc{-0.26} & \hc{+0.77} & \lc{-0.02} & \hc{+2.05} & \hc{+2.12} & \hc{+1.88} & \lc{-0.10} & \hc{+6.63} \\
\textbf{vi} & \lc{-0.99} & \hc{+0.27} & \hc{+0.87} & \lc{-1.70} & \hc{+0.50} & \hc{+2.13} & \hc{+2.47} & \lc{-1.54} & \hc{+7.30} \\
\textbf{ja} & \lc{-1.19} & \lc{-0.44} & \hc{+2.06} & \hc{+1.74} & \hc{+1.54} & \hc{+3.59} & \hc{+2.94} & \lc{-0.01} & \hc{+7.45} \\
\textbf{jw} & \hc{+0.24} & \hc{+3.17} & \hc{+1.87} & \hc{+2.65} & \hc{+4.01} & \hc{+0.04} & \hc{+3.33} & \lc{-0.78} & \hc{+7.00} \\
\textbf{cy} & \lc{-1.50} & \hc{+0.22} & \hc{+0.55} & \hc{+0.78} & \hc{+0.11} & \hc{+0.12} & \lc{-4.17} & \lc{-3.67} & \hc{+0.43} \\
\textbf{mi} & \lc{-3.56} & \lc{-1.53} & \lc{-0.67} & \lc{-1.33} & \lc{-3.13} & \lc{-0.28} & \lc{-1.43} & \lc{-8.49} & \lc{-0.23} \\
\textbf{ti} & \hc{+0.62} & \hc{+1.67} & \hc{+2.46} & \hc{+0.76} & \hc{+3.15} & \hc{+2.71} & \hc{+0.50} & \lc{-2.11} & \hc{+0.03} \\

\bottomrule
\end{tabular}
\caption*{$\Delta$ F1-score (\%)}
\end{minipage}

\vspace{0.3cm}

\begin{minipage}{\textwidth}
\centering
\scriptsize
\textbf{Improvement levels:} 
\colorbox{highperf!30}{\strut~Positive ($>$0)~} 
\colorbox{lowperf!50}{\strut~Zero or Negative ($\leq$\,0)~} 
\end{minipage}

\caption{Per-configuration Pivot NLLB gains ($\Delta$ = \textbf{Pivot NLLB} $-$ \textbf{Zero-shot}) on Gemma 4 across all 81 claim--evidence language pairs.}
\label{fig:nllb_gemma4_delta_heatmap}
\end{figure}

\begin{figure}[H]
\centering
\tiny
\setlength{\tabcolsep}{2.75pt}
\setlength{\fboxsep}{1pt}

\begin{minipage}[t]{0.5\textwidth}
\centering
\renewcommand{\arraystretch}{1.15}
\begin{tabular}{l|ccccccccc}
\toprule
\multirow{2}{*}{\textbf{Claim}} & \multicolumn{9}{c}{\textbf{Evidence Language}} \\
\cmidrule{2-10}
& \textbf{en} & \textbf{es} & \textbf{ru} & \textbf{vi} & \textbf{ja} & \textbf{jw} & \textbf{cy} & \textbf{mi} & \textbf{ti} \\

\midrule
\textbf{en} & \nc{--} & \hc{+1.64} & \hc{+2.78} & \hc{+2.53} & \hc{+0.20} & \hc{+1.68} & \hc{+0.73} & \hc{+1.64} & \hc{+0.69} \\
\textbf{es} & \lc{-0.60} & \lc{-0.30} & \hc{+1.02} & \hc{+0.39} & \lc{-0.92} & \hc{+1.32} & \hc{+0.35} & \hc{+1.37} & \hc{+0.96} \\
\textbf{ru} & \lc{-0.35} & \hc{+1.26} & \hc{+0.69} & \hc{+0.32} & \lc{-0.97} & \lc{-0.18} & \hc{+1.84} & \hc{+0.86} & \hc{+0.57} \\
\textbf{vi} & \hc{+0.67} & \hc{+0.59} & \lc{-0.33} & \hc{+0.24} & \lc{-1.79} & \hc{+0.65} & \hc{+1.11} & \hc{+0.91} & \hc{+0.27} \\
\textbf{ja} & \hc{+0.65} & \hc{+0.48} & \hc{+0.24} & \hc{+0.56} & \lc{-1.43} & \hc{+1.75} & \hc{+1.43} & \hc{+1.68} & \hc{+0.79} \\
\textbf{jw} & \hc{+1.43} & \hc{+0.31} & \lc{-0.03} & \hc{+0.62} & \lc{-1.57} & \hc{+1.48} & \hc{+1.23} & \hc{+1.39} & \hc{+0.03} \\
\textbf{cy} & \hc{+0.09} & \hc{+1.39} & \lc{-0.88} & \hc{+0.08} & \lc{-1.07} & \lc{-1.04} & \hc{+2.00} & \hc{+0.59} & \hc{+0.55} \\
\textbf{mi} & \lc{-0.54} & \hc{+1.30} & \lc{-0.58} & \hc{+0.59} & \lc{-1.47} & \lc{-0.35} & \hc{+0.78} & \hc{+0.67} & \lc{-0.30} \\
\textbf{ti} & \lc{-0.74} & \lc{-0.16} & \lc{-0.38} & \lc{-0.82} & \lc{-2.12} & \lc{-0.64} & \hc{+0.92} & \hc{+0.04} & \hc{+0.82} \\

\bottomrule
\end{tabular}
\caption*{$\Delta$ Accuracy (\%)}
\end{minipage}%
\hfill
%
\begin{minipage}[t]{0.5\textwidth}
\centering
\renewcommand{\arraystretch}{1.15}
\begin{tabular}{l|ccccccccc}
\toprule
\multirow{2}{*}{\textbf{Claim}} & \multicolumn{9}{c}{\textbf{Evidence Language}} \\
\cmidrule{2-10}
& \textbf{en} & \textbf{es} & \textbf{ru} & \textbf{vi} & \textbf{ja} & \textbf{jw} & \textbf{cy} & \textbf{mi} & \textbf{ti} \\

\midrule
\textbf{en} & \nc{--} & \hc{+1.59} & \hc{+2.86} & \hc{+2.66} & \lc{-0.09} & \hc{+2.08} & \hc{+1.71} & \hc{+2.42} & \hc{+1.23} \\
\textbf{es} & \lc{-1.06} & \hc{+0.22} & \hc{+1.17} & \hc{+0.39} & \lc{-1.18} & \hc{+1.72} & \lc{-0.14} & \hc{+1.23} & \lc{-0.15} \\
\textbf{ru} & \lc{-0.36} & \hc{+1.53} & \hc{+1.22} & \hc{+0.43} & \lc{-1.21} & \lc{-0.07} & \hc{+0.88} & \hc{+0.33} & \lc{-1.03} \\
\textbf{vi} & \hc{+0.74} & \hc{+0.78} & \lc{-0.40} & \hc{+0.25} & \lc{-1.93} & \hc{+1.10} & \hc{+0.30} & \hc{+0.66} & \lc{-0.77} \\
\textbf{ja} & \hc{+0.99} & \hc{+1.05} & \hc{+0.43} & \hc{+0.73} & \lc{-1.72} & \hc{+1.92} & \hc{+0.14} & \hc{+0.65} & \lc{-1.18} \\
\textbf{jw} & \hc{+1.54} & \hc{+0.26} & \lc{-0.42} & \hc{+0.29} & \lc{-1.76} & \hc{+2.11} & \lc{-0.35} & \hc{+0.40} & \lc{-2.05} \\
\textbf{cy} & \hc{+0.12} & \hc{+0.88} & \lc{-2.20} & \lc{-0.80} & \lc{-1.71} & \lc{-1.75} & \hc{+2.36} & \lc{-0.19} & \lc{-1.62} \\
\textbf{mi} & \lc{-0.55} & \hc{+1.05} & \lc{-1.30} & \hc{+0.11} & \lc{-1.77} & \lc{-0.59} & \lc{-0.59} & \hc{+0.69} & \lc{-2.49} \\
\textbf{ti} & \lc{-0.73} & \lc{-1.07} & \lc{-1.83} & \lc{-2.38} & \lc{-3.20} & \lc{-2.11} & \lc{-1.05} & \lc{-2.18} & \hc{+0.32} \\
\bottomrule
\end{tabular}
\caption*{$\Delta$ F1-score (\%)}
\end{minipage}

\vspace{0.3cm}

\begin{minipage}{\textwidth}
\centering
\scriptsize
\textbf{Improvement levels:} 
\colorbox{highperf!30}{\strut~Positive ($>$0)~} 
\colorbox{lowperf!50}{\strut~Zero or Negative ($\leq$\,0)~} 
\end{minipage}
\caption{Per-configuration CLAIM gains ($\Delta$ = \textbf{CLAIM} $-$ \textbf{Zero-shot}) on Gemma 4 across all 81 claim--evidence language pairs.}
\label{fig:claim_gemma4_delta_heatmap}
\end{figure}
\begin{figure}[H]
\centering
\tiny
\setlength{\tabcolsep}{2.75pt}
\setlength{\fboxsep}{1pt}

\begin{minipage}[t]{0.5\textwidth}
\centering
\renewcommand{\arraystretch}{1.15}
\begin{tabular}{l|ccccccccc}
\toprule
\multirow{2}{*}{\textbf{Claim}} & \multicolumn{9}{c}{\textbf{Evidence Language}} \\
\cmidrule{2-10}
& \textbf{en} & \textbf{es} & \textbf{ru} & \textbf{vi} & \textbf{ja} & \textbf{jw} & \textbf{cy} & \textbf{mi} & \textbf{ti} \\

\midrule
\textbf{en} & \nc{--} & \hc{+1.27} & \hc{+3.32} & \hc{+5.32} & \hc{+2.70} & \hc{+2.74} & \hc{+3.77} & \hc{+3.81} & \hc{+3.89} \\
\textbf{es} & \hc{+1.88} & \hc{+1.27} & \hc{+2.95} & \hc{+5.08} & \hc{+2.46} & \hc{+2.29} & \hc{+3.93} & \hc{+3.97} & \hc{+3.85} \\
\textbf{ru} & \hc{+1.31} & \hc{+0.98} & \hc{+1.80} & \hc{+4.63} & \hc{+2.21} & \hc{+2.83} & \hc{+3.56} & \hc{+2.38} & \hc{+3.85} \\
\textbf{vi} & \hc{+3.07} & \hc{+3.60} & \hc{+2.91} & \hc{+6.84} & \hc{+2.70} & \hc{+3.07} & \hc{+4.38} & \hc{+3.64} & \hc{+3.52} \\
\textbf{ja} & \hc{+3.07} & \hc{+3.28} & \hc{+2.01} & \hc{+3.85} & \hc{+1.35} & \hc{+3.15} & \hc{+3.40} & \hc{+2.21} & \hc{+2.62} \\
\textbf{jw} & \hc{+2.42} & \hc{+1.97} & \hc{+3.81} & \hc{+5.98} & \hc{+3.97} & \hc{+5.53} & \hc{+6.06} & \hc{+4.87} & \hc{+4.55} \\
\textbf{cy} & \hc{+2.74} & \hc{+3.60} & \hc{+3.15} & \hc{+5.77} & \hc{+3.15} & \hc{+4.18} & \hc{+8.44} & \hc{+4.87} & \hc{+5.24} \\
\textbf{mi} & \hc{+1.39} & \hc{+3.40} & \hc{+1.97} & \hc{+6.27} & \hc{+3.03} & \hc{+3.15} & \hc{+4.71} & \hc{+4.95} & \hc{+5.57} \\
\textbf{ti} & \hc{+1.02} & \hc{+1.19} & \hc{+1.92} & \hc{+2.58} & \hc{+3.48} & \hc{+0.57} & \hc{+1.56} & \hc{+2.01} & \hc{+5.57} \\
\bottomrule
\end{tabular}
\caption*{$\Delta$ Accuracy (\%)}
\end{minipage}%
\hfill
%
\begin{minipage}[t]{0.5\textwidth}
\centering
\renewcommand{\arraystretch}{1.15}
\begin{tabular}{l|ccccccccc}
\toprule
\multirow{2}{*}{\textbf{Claim}} & \multicolumn{9}{c}{\textbf{Evidence Language}} \\
\cmidrule{2-10}
& \textbf{en} & \textbf{es} & \textbf{ru} & \textbf{vi} & \textbf{ja} & \textbf{jw} & \textbf{cy} & \textbf{mi} & \textbf{ti} \\

\midrule
\textbf{en} & \nc{--} & \hc{+2.51} & \hc{+4.36} & \hc{+6.24} & \hc{+3.58} & \hc{+5.18} & \hc{+5.70} & \hc{+5.96} & \hc{+4.50} \\
\textbf{es} & \hc{+2.77} & \hc{+2.13} & \hc{+4.11} & \hc{+5.81} & \hc{+3.74} & \hc{+4.40} & \hc{+5.92} & \hc{+5.98} & \hc{+4.41} \\
\textbf{ru} & \hc{+1.81} & \hc{+2.29} & \hc{+1.87} & \hc{+5.26} & \hc{+2.55} & \hc{+4.77} & \hc{+4.74} & \hc{+3.58} & \hc{+3.62} \\
\textbf{vi} & \hc{+4.87} & \hc{+5.51} & \hc{+4.61} & \hc{+7.62} & \hc{+4.12} & \hc{+4.73} & \hc{+5.18} & \hc{+4.75} & \hc{+2.86} \\
\textbf{ja} & \hc{+3.46} & \hc{+4.22} & \hc{+2.89} & \hc{+4.92} & \hc{+1.60} & \hc{+5.69} & \hc{+5.57} & \hc{+4.41} & \hc{+3.44} \\
\textbf{jw} & \hc{+3.40} & \hc{+3.25} & \hc{+4.79} & \hc{+6.41} & \hc{+5.03} & \hc{+6.44} & \hc{+7.11} & \hc{+5.87} & \hc{+4.20} \\
\textbf{cy} & \hc{+3.71} & \hc{+4.39} & \hc{+4.18} & \hc{+6.40} & \hc{+4.33} & \hc{+5.09} & \hc{+8.65} & \hc{+5.27} & \hc{+4.23} \\
\textbf{mi} & \hc{+1.65} & \hc{+4.13} & \hc{+2.98} & \hc{+6.77} & \hc{+3.60} & \hc{+3.80} & \hc{+5.33} & \hc{+4.92} & \hc{+4.82} \\
\textbf{ti} & \hc{+2.19} & \hc{+2.19} & \hc{+3.01} & \hc{+1.73} & \hc{+4.35} & \lc{-1.14} & \lc{-0.47} & \hc{+0.44} & \hc{+4.00} \\
\bottomrule
\end{tabular}
\caption*{$\Delta$ F1-score (\%)}
\end{minipage}

\vspace{0.3cm}

\begin{minipage}{\textwidth}
\centering
\scriptsize
\textbf{Improvement levels:} 
\colorbox{highperf!30}{\strut~Positive ($>$0)~} 
\colorbox{lowperf!50}{\strut~Zero or Negative ($\leq$\,0)~} 
\end{minipage}

\caption{Per-configuration ORBIT gains ($\Delta$ = \textbf{ORBIT} $-$ \textbf{Zero-shot}) on LFM2-VL across all 81 claim--evidence language pairs.}
\label{fig:orbit_lfm2_delta_heatmap}
\end{figure}

\begin{figure}[H]
\centering
\tiny
\setlength{\tabcolsep}{2.75pt}
\setlength{\fboxsep}{1pt}

\begin{minipage}[t]{0.5\textwidth}
\centering
\renewcommand{\arraystretch}{1.15}
\begin{tabular}{l|ccccccccc}
\toprule
\multirow{2}{*}{\textbf{Claim}} & \multicolumn{9}{c}{\textbf{Evidence Language}} \\
\cmidrule{2-10}
& \textbf{en} & \textbf{es} & \textbf{ru} & \textbf{vi} & \textbf{ja} & \textbf{jw} & \textbf{cy} & \textbf{mi} & \textbf{ti} \\

\midrule
\textbf{en} & \nc{--} & \hc{+2.14} & \hc{+4.22} & \hc{+5.74} & \hc{+5.04} & \hc{+4.47} & \hc{+2.05} & \hc{+3.16} & \hc{+2.05} \\
\textbf{es} & \lc{-1.76} & \lc{-2.91} & \hc{+1.84} & \hc{+5.53} & \hc{+4.18} & \hc{+3.08} & \hc{+1.44} & \hc{+2.26} & \hc{+0.56} \\
\textbf{ru} & \lc{-1.11} & \lc{-0.61} & \lc{-1.02} & \hc{+5.94} & \hc{+3.28} & \hc{+3.69} & \hc{+2.59} & \hc{+3.45} & \hc{+1.14} \\
\textbf{vi} & \lc{-1.27} & \hc{+0.49} & \hc{+2.38} & \hc{+3.04} & \hc{+2.26} & \hc{+3.32} & \hc{+2.63} & \hc{+3.41} & \hc{+1.50} \\
\textbf{ja} & \lc{-2.50} & \lc{-1.72} & \hc{+0.94} & \hc{+4.14} & \lc{-1.51} & \hc{+3.98} & \hc{+3.08} & \hc{+1.93} & \hc{+0.48} \\
\textbf{jw} & \lc{-0.16} & \lc{-0.82} & \hc{+2.42} & \hc{+3.49} & \hc{+2.55} & \hc{+4.31} & \hc{+2.71} & \hc{+3.90} & \hc{+2.69} \\
\textbf{cy} & \lc{-2.62} & \lc{-1.39} & \hc{+0.82} & \hc{+1.85} & \hc{+0.21} & \hc{+2.75} & \hc{+2.67} & \hc{+2.05} & \hc{+3.35} \\
\textbf{mi} & \lc{-3.89} & \lc{-2.83} & \hc{+1.23} & \hc{+2.26} & \lc{-0.77} & \hc{+1.24} & \hc{+1.89} & \hc{+0.83} & \hc{+2.69} \\
\textbf{ti} & \lc{-5.16} & \lc{-5.08} & \lc{-0.70} & \lc{-0.22} & \lc{-2.00} & \hc{+0.01} & \hc{+0.58} & \hc{+1.69} & \hc{+4.66} \\

\bottomrule
\end{tabular}
\caption*{$\Delta$ Accuracy (\%)}
\end{minipage}%
\hfill
%
\begin{minipage}[t]{0.5\textwidth}
\centering
\renewcommand{\arraystretch}{1.15}
\begin{tabular}{l|ccccccccc}
\toprule
\multirow{2}{*}{\textbf{Claim}} & \multicolumn{9}{c}{\textbf{Evidence Language}} \\
\cmidrule{2-10}
& \textbf{en} & \textbf{es} & \textbf{ru} & \textbf{vi} & \textbf{ja} & \textbf{jw} & \textbf{cy} & \textbf{mi} & \textbf{ti} \\

\midrule
\textbf{en} & \nc{--} & \hc{+4.67} & \hc{+6.54} & \hc{+8.67} & \hc{+7.69} & \hc{+6.68} & \hc{+1.65} & \hc{+2.92} & \lc{-2.94} \\
\textbf{es} & \lc{-2.53} & \lc{-3.54} & \hc{+1.14} & \hc{+7.99} & \hc{+7.29} & \hc{+4.70} & \hc{+1.05} & \hc{+1.79} & \lc{-4.02} \\
\textbf{ru} & \lc{-2.22} & \lc{-1.10} & \lc{-2.91} & \hc{+8.50} & \hc{+5.80} & \hc{+4.66} & \hc{+1.93} & \hc{+2.79} & \lc{-3.03} \\
\textbf{vi} & \lc{-1.97} & \hc{+0.26} & \hc{+2.27} & \hc{+4.68} & \hc{+5.54} & \hc{+4.98} & \hc{+3.03} & \hc{+3.11} & \lc{-2.50} \\
\textbf{ja} & \lc{-4.26} & \lc{-2.80} & \lc{-0.40} & \hc{+6.81} & \hc{+0.06} & \hc{+4.85} & \hc{+2.14} & \hc{+1.07} & \lc{-3.49} \\
\textbf{jw} & \lc{-0.78} & \lc{-1.37} & \hc{+1.69} & \hc{+5.77} & \hc{+5.72} & \hc{+4.79} & \hc{+2.66} & \hc{+3.25} & \lc{-1.20} \\
\textbf{cy} & \lc{-5.16} & \lc{-3.12} & \lc{-0.94} & \hc{+3.16} & \hc{+1.71} & \hc{+3.20} & \hc{+0.99} & \hc{+1.11} & \lc{-1.12} \\
\textbf{mi} & \lc{-6.45} & \lc{-4.75} & \lc{-0.84} & \hc{+3.24} & \hc{+0.03} & \hc{+0.28} & \hc{+0.65} & \lc{-1.62} & \lc{-2.60} \\
\textbf{ti} & \lc{-9.89} & \lc{-9.57} & \lc{-4.66} & \lc{-1.79} & \lc{-4.80} & \lc{-3.65} & \lc{-2.70} & \lc{-1.86} & \hc{+0.05} \\
\bottomrule
\end{tabular}
\caption*{$\Delta$ F1-score (\%)}
\end{minipage}

\vspace{0.3cm}

\begin{minipage}{\textwidth}
\centering
\scriptsize
\textbf{Improvement levels:} 
\colorbox{highperf!30}{\strut~Positive ($>$0)~} 
\colorbox{lowperf!50}{\strut~Zero or Negative ($\leq$\,0)~} 
\end{minipage}

\caption{Per-configuration Self-translate CoT gains ($\Delta$ = \textbf{Self-translate CoT} $-$ \textbf{Zero-shot}) on LFM2-VL across all 81 claim--evidence language pairs.}
\label{fig:cot_lfm2_delta_heatmap}
\end{figure}

\begin{figure}[H]
\centering
\tiny
\setlength{\tabcolsep}{2.75pt}
\setlength{\fboxsep}{1pt}

\begin{minipage}[t]{0.5\textwidth}
\centering
\renewcommand{\arraystretch}{1.15}
\begin{tabular}{l|ccccccccc}
\toprule
\multirow{2}{*}{\textbf{Claim}} & \multicolumn{9}{c}{\textbf{Evidence Language}} \\
\cmidrule{2-10}
& \textbf{en} & \textbf{es} & \textbf{ru} & \textbf{vi} & \textbf{ja} & \textbf{jw} & \textbf{cy} & \textbf{mi} & \textbf{ti} \\

\midrule
\textbf{en} & \nc{--} & \hc{+0.25} & \hc{+2.83} & \hc{+5.82} & \hc{+3.20} & \hc{+3.93} & \hc{+1.36} & \hc{+1.40} & \hc{+6.02} \\
\textbf{es} & \lc{-0.25} & \lc{-2.95} & \hc{+2.83} & \hc{+5.25} & \hc{+1.36} & \hc{+2.29} & \hc{+0.46} & \hc{+0.46} & \hc{+5.07} \\
\textbf{ru} & \hc{+1.56} & \hc{+0.29} & \hc{+0.42} & \hc{+5.66} & \hc{+2.38} & \hc{+3.60} & \hc{+2.05} & \hc{+2.38} & \hc{+6.38} \\
\textbf{vi} & \hc{+2.13} & \hc{+3.36} & \hc{+5.37} & \hc{+3.28} & \hc{+3.73} & \hc{+3.72} & \hc{+2.01} & \hc{+2.46} & \hc{+6.02} \\
\textbf{ja} & \lc{-0.33} & \lc{-1.64} & \hc{+1.89} & \hc{+3.53} & \lc{-2.94} & \hc{+1.80} & \hc{+0.83} & \hc{+1.69} & \hc{+4.54} \\
\textbf{jw} & \hc{+3.24} & \hc{+0.98} & \hc{+4.84} & \hc{+3.81} & \hc{+3.69} & \hc{+2.90} & \hc{+1.81} & \hc{+2.55} & \hc{+6.18} \\
\textbf{cy} & \lc{-0.04} & \hc{+0.45} & \hc{+1.11} & \hc{+1.52} & \hc{+0.17} & \hc{+2.00} & \hc{+1.15} & \hc{+0.74} & \hc{+4.83} \\
\textbf{mi} & \lc{-2.66} & \lc{-1.52} & \hc{+0.66} & \hc{+1.19} & \lc{-1.59} & \hc{+0.32} & \lc{-0.65} & \lc{-0.65} & \hc{+3.56} \\
\textbf{ti} & \hc{+1.55} & \hc{+1.55} & \hc{+5.11} & \hc{+3.84} & \hc{+2.58} & \hc{+2.61} & \hc{+1.15} & \hc{+1.52} & \hc{+7.41} \\
\bottomrule
\end{tabular}
\caption*{$\Delta$ Accuracy (\%)}
\end{minipage}%
\hfill
%
\begin{minipage}[t]{0.5\textwidth}
\centering
\renewcommand{\arraystretch}{1.15}
\begin{tabular}{l|ccccccccc}
\toprule
\multirow{2}{*}{\textbf{Claim}} & \multicolumn{9}{c}{\textbf{Evidence Language}} \\
\cmidrule{2-10}
& \textbf{en} & \textbf{es} & \textbf{ru} & \textbf{vi} & \textbf{ja} & \textbf{jw} & \textbf{cy} & \textbf{mi} & \textbf{ti} \\

\midrule
\textbf{en} & \nc{--} & \hc{+0.44} & \hc{+2.64} & \hc{+7.08} & \hc{+3.04} & \hc{+5.72} & \hc{+1.15} & \hc{+0.02} & \hc{+5.14} \\
\textbf{es} & \lc{-0.05} & \lc{-2.88} & \hc{+2.86} & \hc{+6.15} & \hc{+1.49} & \hc{+3.51} & \lc{-0.05} & \lc{-1.56} & \hc{+3.57} \\
\textbf{ru} & \hc{+1.53} & \hc{+0.67} & \lc{-0.30} & \hc{+7.02} & \hc{+1.96} & \hc{+5.07} & \hc{+1.40} & \hc{+0.64} & \hc{+4.98} \\
\textbf{vi} & \hc{+2.80} & \hc{+4.43} & \hc{+6.45} & \hc{+3.86} & \hc{+4.61} & \hc{+5.91} & \hc{+2.67} & \hc{+1.99} & \hc{+5.52} \\
\textbf{ja} & \lc{-1.31} & \lc{-2.41} & \hc{+1.11} & \hc{+4.18} & \lc{-4.70} & \hc{+2.35} & \lc{-0.27} & \lc{-0.76} & \hc{+3.21} \\
\textbf{jw} & \hc{+4.76} & \hc{+2.73} & \hc{+6.59} & \hc{+5.82} & \hc{+4.80} & \hc{+3.51} & \hc{+1.48} & \hc{+1.15} & \hc{+4.56} \\
\textbf{cy} & \lc{-0.76} & \lc{-0.63} & \hc{+0.73} & \hc{+2.10} & \lc{-1.01} & \hc{+2.31} & \lc{-0.91} & \lc{-1.86} & \hc{+1.62} \\
\textbf{mi} & \lc{-3.24} & \lc{-2.36} & \hc{+0.52} & \hc{+1.60} & \lc{-2.56} & \lc{-0.26} & \lc{-2.09} & \lc{-4.24} & \lc{-0.05} \\
\textbf{ti} & \hc{+3.04} & \hc{+3.06} & \hc{+7.48} & \hc{+6.48} & \hc{+3.48} & \hc{+5.65} & \hc{+2.65} & \hc{+1.70} & \hc{+7.04} \\
\bottomrule
\end{tabular}
\caption*{$\Delta$ F1-score (\%)}
\end{minipage}

\vspace{0.3cm}

\begin{minipage}{\textwidth}
\centering
\scriptsize
\textbf{Improvement levels:} 
\colorbox{highperf!30}{\strut~Positive ($>$0)~} 
\colorbox{lowperf!50}{\strut~Zero or Negative ($\leq$\,0)~} 
\end{minipage}

\caption{Per-configuration Pivot NLLB gains ($\Delta$ = \textbf{Pivot NLLB} $-$ \textbf{Zero-shot}) on LFM2-VL across all 81 claim--evidence language pairs.}
\label{fig:nllb_lfm2_delta_heatmap}
\end{figure}

\begin{figure}[H]
\centering
\tiny
\setlength{\tabcolsep}{2.75pt}
\setlength{\fboxsep}{1pt}

\begin{minipage}[t]{0.5\textwidth}
\centering
\renewcommand{\arraystretch}{1.15}
\begin{tabular}{l|ccccccccc}
\toprule
\multirow{2}{*}{\textbf{Claim}} & \multicolumn{9}{c}{\textbf{Evidence Language}} \\
\cmidrule{2-10}
& \textbf{en} & \textbf{es} & \textbf{ru} & \textbf{vi} & \textbf{ja} & \textbf{jw} & \textbf{cy} & \textbf{mi} & \textbf{ti} \\

\midrule
\textbf{en} & \nc{--} & \lc{-2.30} & \hc{+1.35} & \hc{+7.04} & \hc{+0.61} & \hc{+2.97} & \hc{+4.79} & \hc{+4.85} & \hc{+4.87} \\
\textbf{es} & \lc{-3.33} & \lc{-3.81} & \hc{+0.45} & \hc{+5.00} & \hc{+0.81} & \hc{+1.54} & \hc{+2.91} & \hc{+4.83} & \hc{+5.07} \\
\textbf{ru} & \lc{-2.59} & \lc{-2.54} & \lc{-0.33} & \hc{+5.15} & \lc{-1.40} & \hc{+0.55} & \hc{+2.25} & \hc{+3.35} & \hc{+3.43} \\
\textbf{vi} & \lc{-1.27} & \hc{+0.32} & \hc{+1.71} & \hc{+5.15} & \hc{+0.69} & \hc{+2.85} & \hc{+2.21} & \hc{+4.58} & \hc{+4.11} \\
\textbf{ja} & \lc{-1.19} & \lc{-2.54} & \lc{-0.66} & \hc{+4.17} & \lc{-0.91} & \hc{+1.28} & \hc{+1.43} & \hc{+2.33} & \lc{-0.25} \\
\textbf{jw} & \lc{-2.34} & \lc{-2.42} & \hc{+0.70} & \hc{+3.76} & \hc{+0.41} & \hc{+1.62} & \hc{+3.61} & \hc{+4.58} & \hc{+4.10} \\
\textbf{cy} & \lc{-1.95} & \lc{-1.74} & \lc{-0.63} & \hc{+3.67} & \hc{+0.06} & \hc{+1.00} & \hc{+3.65} & \hc{+0.86} & \hc{+5.42} \\
\textbf{mi} & \lc{-2.54} & \lc{-2.18} & \hc{+0.49} & \hc{+4.71} & \hc{+0.20} & \hc{+0.51} & \hc{+1.31} & \hc{+2.29} & \hc{+4.34} \\
\textbf{ti} & \lc{-1.06} & \lc{-1.71} & \lc{-1.02} & \hc{+3.65} & \hc{+1.24} & \lc{-1.55} & \hc{+0.37} & \hc{+0.66} & \hc{+7.69} \\
\bottomrule
\end{tabular}
\caption*{$\Delta$ Accuracy (\%)}
\end{minipage}%
\hfill
%
\begin{minipage}[t]{0.5\textwidth}
\centering
\renewcommand{\arraystretch}{1.15}
\begin{tabular}{l|ccccccccc}
\toprule
\multirow{2}{*}{\textbf{Claim}} & \multicolumn{9}{c}{\textbf{Evidence Language}} \\
\cmidrule{2-10}
& \textbf{en} & \textbf{es} & \textbf{ru} & \textbf{vi} & \textbf{ja} & \textbf{jw} & \textbf{cy} & \textbf{mi} & \textbf{ti} \\

\midrule
\textbf{en} & \nc{--} & \lc{-4.96} & \lc{-1.10} & \hc{+5.42} & \lc{-0.79} & \hc{+3.17} & \hc{+6.08} & \hc{+4.93} & \hc{+4.13} \\
\textbf{es} & \lc{-5.96} & \lc{-6.51} & \lc{-2.39} & \hc{+3.07} & \lc{-0.69} & \hc{+1.05} & \hc{+4.20} & \hc{+3.83} & \hc{+4.05} \\
\textbf{ru} & \lc{-5.21} & \lc{-5.08} & \lc{-3.64} & \hc{+3.23} & \lc{-3.29} & \hc{+0.31} & \hc{+3.57} & \hc{+2.59} & \hc{+2.15} \\
\textbf{vi} & \lc{-4.33} & \lc{-1.85} & \hc{+0.40} & \hc{+2.23} & \lc{-0.42} & \hc{+3.08} & \hc{+4.28} & \hc{+4.22} & \hc{+3.85} \\
\textbf{ja} & \lc{-3.31} & \lc{-4.64} & \lc{-2.20} & \hc{+3.27} & \lc{-2.13} & \hc{+1.55} & \hc{+3.58} & \hc{+2.49} & \hc{+1.54} \\
\textbf{jw} & \lc{-5.88} & \lc{-5.15} & \lc{-1.36} & \hc{+1.47} & \lc{-1.55} & \lc{-0.79} & \hc{+3.99} & \hc{+2.95} & \hc{+6.07} \\
\textbf{cy} & \lc{-5.09} & \lc{-3.92} & \lc{-1.18} & \hc{+2.98} & \lc{-0.58} & \hc{+1.01} & \hc{+2.63} & \hc{+1.38} & \hc{+4.31} \\
\textbf{mi} & \lc{-6.41} & \lc{-4.90} & \lc{-1.26} & \hc{+2.96} & \lc{-1.26} & \lc{-0.22} & \hc{+1.82} & \lc{-1.15} & \hc{+6.11} \\
\textbf{ti} & \lc{-1.36} & \lc{-1.34} & \hc{+1.14} & \hc{+3.87} & \hc{+1.38} & \hc{+1.05} & \hc{+3.88} & \hc{+4.02} & \hc{+6.57} \\
\bottomrule
\end{tabular}
\caption*{$\Delta$ F1-score (\%)}
\end{minipage}

\vspace{0.3cm}

\begin{minipage}{\textwidth}
\centering
\scriptsize
\textbf{Improvement levels:} 
\colorbox{highperf!30}{\strut~Positive ($>$0)~} 
\colorbox{lowperf!50}{\strut~Zero or Negative ($\leq$\,0)~} 
\end{minipage}
\caption{Per-configuration CLAIM gains ($\Delta$ = \textbf{CLAIM} $-$ \textbf{Zero-shot}) on LFM2-VL across all 81 claim--evidence language pairs.}
\label{fig:claim_lfm2_delta_heatmap}
\end{figure}

\begin{table}[H]
\centering
\setlength{\tabcolsep}{6.75pt}
\renewcommand{\arraystretch}{1.15}
\begin{tabular}{l|ccccccccc}
\toprule
\multirow{2}{*}{\textbf{Claim}} & \multicolumn{9}{c}{\textbf{Evidence Language}} \\
\cmidrule{2-10}
& \textbf{en} & \textbf{es} & \textbf{ru} & \textbf{vi} & \textbf{ja} & \textbf{jw} & \textbf{cy} & \textbf{mi} & \textbf{ti} \\
\midrule
\textbf{en} & \nc{--} & \hc{+2.14} & \hc{+3.24} & \hc{+3.68} & \hc{+3.70} & \hc{+3.51} & \hc{+2.86} & \hc{+3.53} & \hc{+1.71} \\
\textbf{es} & \hc{+0.76} & \hc{+1.02} & \hc{+3.16} & \hc{+2.40} & \hc{+2.36} & \hc{+3.53} & \hc{+2.82} & \hc{+1.65} & \hc{+1.71} \\
\textbf{ru} & \hc{+0.88} & \hc{+1.75} & \hc{+2.77} & \hc{+2.40} & \hc{+3.76} & \hc{+4.64} & \hc{+3.73} & \hc{+1.61} & \hc{+2.19} \\
\textbf{vi} & \hc{+1.50} & \hc{+3.25} & \hc{+3.84} & \hc{+4.15} & \hc{+4.04} & \hc{+4.77} & \hc{+3.30} & \hc{+4.46} & \hc{+1.70} \\
\textbf{ja} & \hc{+0.49} & \hc{+1.39} & \hc{+3.45} & \hc{+1.18} & \hc{+4.38} & \hc{+3.01} & \hc{+3.23} & \hc{+0.64} & \hc{+1.66} \\
\textbf{jw} & \hc{+2.96} & \hc{+2.82} & \hc{+2.65} & \hc{+3.15} & \hc{+3.40} & \hc{+3.52} & \hc{+2.45} & \hc{+2.07} & \hc{+0.93} \\
\textbf{cy} & \hc{+3.32} & \hc{+1.48} & \hc{+3.14} & \hc{+2.42} & \hc{+3.23} & \hc{+4.15} & \hc{+1.67} & \hc{+0.12} & \hc{+1.29} \\
\textbf{mi} & \hc{+0.43} & \lc{-0.54} & \lc{-1.01} & \lc{-0.87} & \lc{-1.51} & \lc{-1.04} & \hc{+0.13} & \hc{+0.35} & \lc{-0.53} \\
\textbf{ti} & \hc{+1.38} & \lc{-1.54} & \lc{-3.51} & \lc{-1.10} & \lc{-2.34} & \lc{-1.62} & \lc{-1.58} & \lc{-2.08} & \hc{+1.95} \\
\bottomrule
\end{tabular}
\caption{ORBIT $\Delta$F1-score on Qwen3-VL for \textit{Supported} class relative to zero-shot baseline.}
\label{tab:supported_ORBIT_heatmap}
\end{table}

\begin{table}[H]
\centering
\setlength{\tabcolsep}{6pt}
\renewcommand{\arraystretch}{1.15}

\begin{tabular}{l|ccccccccc}
\toprule
\multirow{2}{*}{\textbf{Claim}} & \multicolumn{9}{c}{\textbf{Evidence Language}} \\
\cmidrule{2-10}
& \textbf{en} & \textbf{es} & \textbf{ru} & \textbf{vi} & \textbf{ja} & \textbf{jw} & \textbf{cy} & \textbf{mi} & \textbf{ti} \\
\midrule
\textbf{en} & \nc{--} & \hc{+0.86} & \hc{+2.53} & \hc{+2.19} & \hc{+4.35} & \hc{+6.55} & \hc{+9.04} & \hc{+9.81} & \hc{+9.33} \\
\textbf{es} & \hc{+0.72} & \hc{+1.59} & \hc{+1.28} & \hc{+2.73} & \hc{+4.29} & \hc{+7.33} & \hc{+7.26} & \hc{+9.01} & \hc{+8.84} \\
\textbf{ru} & \hc{+0.00} & \hc{+1.80} & \hc{+0.79} & \hc{+2.49} & \hc{+4.53} & \hc{+9.84} & \hc{+8.63} & \hc{+8.46} & \hc{+6.82} \\
\textbf{vi} & \lc{-0.28} & \hc{+1.54} & \hc{+1.27} & \hc{+2.28} & \hc{+5.13} & \hc{+8.09} & \hc{+7.71} & \hc{+11.30} & \hc{+7.63} \\
\textbf{ja} & \hc{+0.01} & \hc{+1.60} & \hc{+1.54} & \hc{+2.96} & \hc{+4.44} & \hc{+8.26} & \hc{+9.15} & \hc{+10.90} & \hc{+7.25} \\
\textbf{jw} & \hc{+2.85} & \hc{+4.23} & \hc{+5.84} & \hc{+6.08} & \hc{+8.56} & \hc{+6.54} & \hc{+8.81} & \hc{+12.79} & \hc{+8.70} \\
\textbf{cy} & \hc{+5.04} & \hc{+5.54} & \hc{+5.81} & \hc{+5.80} & \hc{+8.88} & \hc{+11.19} & \hc{+9.02} & \hc{+7.86} & \hc{+2.81} \\
\textbf{mi} & \hc{+3.75} & \hc{+3.70} & \hc{+4.60} & \hc{+4.06} & \hc{+7.33} & \hc{+8.20} & \hc{+5.40} & \hc{+7.99} & \hc{+1.12} \\
\textbf{ti} & \hc{+3.99} & \hc{+3.22} & \hc{+3.30} & \hc{+3.90} & \hc{+5.51} & \hc{+4.75} & \hc{+1.14} & \hc{+1.75} & \hc{+10.98} \\
\bottomrule
\end{tabular}
\caption{ORBIT $\Delta$F1-score on Qwen3-VL for \textit{Refuted} class relative to zero-shot baseline.}
\label{tab:Refuted_ORBIT_heatmap}
\end{table}

\begin{table}[H]
\centering
\setlength{\tabcolsep}{6.25pt}
\renewcommand{\arraystretch}{1.15}

\begin{tabular}{l|ccccccccc}
\toprule
\multirow{2}{*}{\textbf{Claim}} & \multicolumn{9}{c}{\textbf{Evidence Language}} \\
\cmidrule{2-10}
& \textbf{en} & \textbf{es} & \textbf{ru} & \textbf{vi} & \textbf{ja} & \textbf{jw} & \textbf{cy} & \textbf{mi} & \textbf{ti} \\
\midrule
\textbf{en} & \nc{--} & \lc{-9.54} & \lc{-8.02} & \lc{-8.49} & \lc{-11.16} & \lc{-7.36} & \lc{-4.80} & \lc{-3.00} & \lc{-0.06} \\
\textbf{es} & \lc{-9.97} & \lc{-10.51} & \lc{-8.38} & \lc{-9.42} & \lc{-11.79} & \lc{-5.82} & \lc{-5.20} & \lc{-3.80} & \hc{+0.92} \\
\textbf{ru} & \lc{-10.69} & \lc{-10.73} & \lc{-9.91} & \lc{-9.62} & \lc{-10.32} & \lc{-8.21} & \lc{-5.57} & \lc{-4.01} & \lc{-0.22} \\
\textbf{vi} & \lc{-11.56} & \lc{-9.19} & \lc{-9.30} & \lc{-9.61} & \lc{-8.99} & \lc{-6.86} & \lc{-4.79} & \lc{-2.77} & \lc{-0.42} \\
\textbf{ja} & \lc{-12.36} & \lc{-9.90} & \lc{-9.24} & \lc{-10.28} & \lc{-11.81} & \lc{-8.36} & \lc{-5.38} & \lc{-4.94} & \lc{-1.38} \\
\textbf{jw} & \lc{-10.22} & \lc{-6.90} & \lc{-5.97} & \lc{-8.05} & \lc{-6.24} & \lc{-8.15} & \lc{-4.03} & \lc{-2.26} & \lc{-0.16} \\
\textbf{cy} & \lc{-6.45} & \lc{-4.26} & \lc{-5.08} & \lc{-4.65} & \lc{-4.58} & \lc{-2.04} & \lc{-5.22} & \lc{-0.53} & \lc{-0.49} \\
\textbf{mi} & \lc{-3.02} & \lc{-2.53} & \lc{-3.10} & \lc{-1.76} & \lc{-3.39} & \lc{-2.60} & \lc{-0.65} & \lc{-2.26} & \hc{+0.10} \\
\textbf{ti} & \lc{-2.77} & \lc{-1.68} & \hc{+0.06} & \lc{-0.02} & \lc{-1.55} & \hc{+0.19} & \hc{+0.64} & \hc{+0.47} & \lc{-2.13} \\
\bottomrule
\end{tabular}
\caption{ORBIT $\Delta$F1-score on Qwen3-VL for \textit{NEI} class relative to zero-shot baseline.}
\label{tab:NEI_ORBIT_heatmap}
\end{table}

\end{document}